%% file: main.tex
\pdfoutput=1
\documentclass[10pt,twocolumn,letterpaper]{article}

\usepackage[pagenumbers]{cvpr} 

\usepackage{graphicx}
\usepackage[accsupp]{axessibility}
\usepackage[export]{adjustbox}
\usepackage{soul}
\usepackage{amsmath}
\usepackage{amssymb}
\usepackage{booktabs}
\usepackage{pifont}
\usepackage{color}
\usepackage{xcolor}
\usepackage{bm}
\usepackage{multirow}
\usepackage{diagbox}
\usepackage{soul}
\usepackage{arydshln}
\usepackage{bbm}
\usepackage{subcaption}
\usepackage{wrapfig}
\usepackage{makecell}
\usepackage{xurl}
\usepackage[bottom,flushmargin,hang,multiple]{footmisc}

\usepackage[subtle]{savetrees}

\definecolor{citecolor}{HTML}{0071bc}

\usepackage[pagebackref=true,breaklinks=true,colorlinks,citecolor=citecolor,bookmarks=false]{hyperref}

\usepackage[capitalize]{cleveref}
\crefname{section}{Sec.}{Secs.}
\Crefname{section}{Section}{main}
\Crefname{table}{Table}{Tables}
\crefname{table}{Tab.}{Tabs.}
\Crefname{figure}{Figure}{Figures}
\crefname{figure}{Fig.}{Figs.}

\newcommand\blfootnote[1]{%
  \begingroup
  \renewcommand\thefootnote{}\footnote{#1}%
  \addtocounter{footnote}{-1}%
  \endgroup
}

\def\cvprPaperID{11960} 
\def\confName{CVPR}
\def\confYear{2023}

\begin{document}



\title{Unbiased Scene Graph Generation in Videos \vspace{-1em}}
\author{Sayak Nag$^1$, Kyle Min$^2$, Subarna Tripathi$^2$, Amit K. Roy-Chowdhury$^1$\\
$^1$University of California, Riverside, USA, $^2$Intel Corporation, USA \\
\small{\texttt{snag005@ucr.edu}, \texttt{kyle.min@intel.com}, \texttt{subarna.tripathi@intel.com}, \texttt{amitrc@ece.ucr.edu}
}}



\maketitle


\input{0_abstract.tex}

\input{1_introduction.tex}

\input{2_related_works.tex}

\input{3_method.tex}

\input{4_experiments.tex}

\input{5_conclusion.tex}

{\small
\bibliographystyle{ieee_fullname}
\bibliography{egbib}
}
\clearpage

\section*{SUPPLEMENTARY MATERIAL}
\input{Supp}

\end{document}

%% file: 0_abstract.tex
\begin{abstract}
The task of dynamic scene graph generation (SGG) from videos is complicated and challenging due to the inherent dynamics of a scene, temporal fluctuation of model predictions, and the long-tailed distribution of the visual relationships 
in addition to the already existing challenges in image-based SGG. Existing methods for dynamic SGG have primarily
focused on capturing spatio-temporal context using complex
architectures without addressing the challenges mentioned above, especially the long-tailed distribution of relationships. This often leads to the generation of biased scene graphs. To address these challenges, we introduce a new framework called TEMPURA: TEmporal consistency and Memory Prototype guided UnceRtainty Attenuation for unbiased dynamic SGG. TEMPURA employs object-level temporal consistencies via transformer-based sequence modeling, learns to synthesize unbiased relationship representations using memory-guided training, and attenuates the predictive uncertainty of visual relations using a Gaussian Mixture Model (GMM).  Extensive experiments demonstrate that our method achieves significant (up to 10\% in some cases) performance gain over existing methods highlighting its superiority in generating more unbiased scene graphs. \\
Code: \href{https://github.com/sayaknag/unbiasedSGG.git}{https://github.com/sayaknag/unbiasedSGG.git}
\blfootnote{Accepted for publication to IEEE/CVF Conference on Computer Vision and Pattern Recognition (CVPR) 2023 }
\end{abstract}

%% file: 1_introduction.tex
\section{Introduction}
Scene graphs provide a holistic scene understanding that can bridge the gap between vision and language \cite{krishnavisualgenome,Action_genome_Ji_2020_CVPR}. This has made image scene graphs very popular for high-level reasoning tasks such as captioning \cite{Li2017SceneGG,10.1145/3195106.3195114}, image retrieval \cite{wang2020cross,schuster-etal-2015-generating}, human-object interaction (HOI) \cite{liu2020forecasting}, and visual question answering (VQA)\cite{GQA_Hudson_2019_CVPR}. 
Although significant strides have been made in scene graph generation (SGG) from static images \cite{krishnavisualgenome, RelDN,neural_motifs_2017,Li2017SceneGG,VCTree_Tang_2019_CVPR,zhang2019vrd,lu2016visual,lin2020gps,li2022sgtr,desai2021learning}, research on dynamic SGG is still in its nascent stage.

Dynamic SGG involves grounding visual relationships jointly in space and time. It is aimed at obtaining a structured representation of a scene in each video frame along with learning the temporal evolution of the relationships between each pair of objects. Such a detailed and structured form of video understanding is akin to how humans perceive real-world activities \cite{michotte2017perception,baradel2018object,Action_genome_Ji_2020_CVPR} and
with the exponential growth of video data, it is necessary to make similar strides in dynamic SGG.


\input{long_tail_intro.tex}

In recent years, a handful of works have attempted to address dynamic SGG \cite{teng2021target,STTran_2021,ji_HORT,APT_Li_2022_CVPR,simple_baseline_bmvc}, with a majority of them leveraging the superior sequence processing capability of transformers \cite{vaswani2017attention,arnab2021vivit,han2022survey,carion2020end,nawhal2021activity,ju2022prompting,videoBERT_Sun_2019_ICCV}. 
These methods simply focused on designing more complex models to aggregate spatio-temporal contextual information in a video but fail to address the data imbalance of the relationship/predicate classes, and although their performance is encouraging under the Recall@k metric, this metric is biased toward the highly populated classes. An alternative metric was proposed in \cite{KERN,VCTree_Tang_2019_CVPR} called mean-Recall@k which quantifies how SGG models perform over all the classes and not just the high-frequent ones.

Fig \ref{subfig:long_tail} shows the long-tailed distribution of predicate classes in the benchmark Action Genome~\cite{Action_genome_Ji_2020_CVPR} dataset and Fig \ref{subfig:sota_behaviour} highlights the failure of some existing state-of-the-art methods is classifying the relationships/predicates in the tail of the distribution.
The high recall values in prior works suggest that they may have a tendency to overfit on popular predicate classes (e.g. \emph{in front of / not looking at}), without considering how the performances on rare classes (e.g. \emph{eating/wiping}) are getting impacted \cite{desai2021learning}.
Predicates lying in the tails often provide more informative depictions of underlying actions and activities in the video. 
Thus, it is important to be able to measure a model's long-term performance not only on the frequently occurring relationships but also on the infrequently occurring ones.

Data imbalance is, however, not the only challenge in dynamic SGG. As shown in Fig.~\ref{fig:noisy_anno} and Fig.~\ref{fig:inconsistent_obj_clf}, several other factors, including noisy annotations, motion blur, temporal fluctuations of predictions, and a need to focus on only \emph{active} objects that are involved in an action contribute to the bias in training dynamic SGG models \cite{estimating_bias2021}. As a result, the visual relationship predictions have high uncertainty, thereby increasing the challenge of dynamic SGG manyfold.

\input{annotation_characteristics.tex}

In this paper, we address these sources of bias in dynamic SGG and propose methods to compensate for them. We identify missing annotations, multi-label mapping, and triplet ($<subject-predicate-object>$) variability  (Fig \ref{fig:noisy_anno}) as labeling noise, which coupled with the inherent temporal fluctuations in a video can be attributed as data noise that can be modeled as the \emph{aleatoric uncertainty}\cite{der2009aleatory}. Another form of uncertainty called the \emph{epistemic uncertainty}, relates to misleading model predictions due to a lack of sufficient observations \cite{Kendall_uncertainty} and is more prevalent for long-tailed data \cite{imbalanced_uncertainty}. To address the bias in training SGG models \cite{estimating_bias2021} and generate more unbiased scene graphs, it is necessary to model and attenuate the predictive uncertainty of an SGG model. While multi-model deep ensembles \cite{dropout,lakshminarayanan2017simple} can be effective, they are computationally expensive for large-scale video understanding. Therefore, we employ the concepts of single model uncertainty based on Mixture Density Networks (MDN) \cite{tagasovska2019single,Kendall_uncertainty,choi2018uncertainty} and design the predicate classification head as a Gaussian Mixture Model (GMM) \cite{choi2018uncertainty,choi2021active}. The GMM-based predicate classification loss penalizes the model if the predictive uncertainty of a sample is high, thereby, attenuating the effects of noisy SGG annotations.


Due to the long-tailed bias of SGG datasets, the predicate embeddings learned by existing dynamic SGG frameworks significantly underfit to the data-poor classes. Since each object pair can have multiple correct predicates (Fig \ref{fig:noisy_anno}), many relationship classes share similar visual characteristics. Exploiting this factor, we propose a memory-guided training strategy to debias the predicate embeddings by facilitating knowledge transfer from the data-rich to the data-poor classes sharing similar characteristics. This approach is inspired by recent advances in meta-learning and memory-guided training for low-shot, and long-tail image recognition \cite{parisot2022long,NTM,mem-aug-meta,snell2017prototypical,zhu2020inflated}, whereby a memory bank, composed of a set of prototypical abstractions \cite{snell2017prototypical} each compressing information about a predicate class, is designed. We propose a progressive memory computation approach and an attention-based information diffusion strategy \cite{vaswani2017attention}. Backpropagating while using this approach, teaches the model to \emph{learn how to generate} more balanced predicate representations generalizable to all the classes. 

\input{inconsitent_object_clf.tex}

Finally, to ensure the correctness of a generated graph, accurate classification of both nodes (objects) and edges (predicates) is crucial. While existing dynamic SGG methods focus on innovative visual relation classification \cite{STTran_2021,ji_HORT,APT_Li_2022_CVPR}, object classification is typically based on proposals from off-the-shelf object detectors \cite{ren_frcnn}. These detectors may fail to compensate for dynamic nuances in videos such as motion blur, abrupt background changes, occlusion, etc. leading to inconsistent object classification. While some works use bulky tracking algorithms to address this issue \cite{teng2021target}, we propose a simpler yet highly effective learning-based approach combining the superior sequence processing capability of transformers \cite{vaswani2017attention}, with the discriminative power of contrastive learning \cite{hadsell2006dimensionality} to ensure more temporally consistent object classification.
Therefore, combining the principles of temporally consistent object detection, uncertainty-aware learning, and memory-guided training, we design our framework called \textbf{TEMPURA}: \textbf{TE}mporal consistency and \textbf{M}emory \textbf{P}rototype guided \textbf{U}nce\textbf{R}tainty \textbf{A}tentuation for unbiased dynamic SGG.
To the best of our knowledge, \emph{this is the first study that explicitly addresses all sources of bias in dynamic scene graph generation.}

The major contributions of this paper are: 1) TEMPURA models the predictive uncertainty associated with dynamic SGG and attenuates the effect of noisy annotations to produce more unbiased scene graphs. 2) Utilizing a novel memory-guided training approach, TEMPURA learns to generate more unbiased predicate representations by diffusing knowledge from highly frequent predicate classes to rare ones. 3) Utilizing a transformer-based sequence processing mechanism, TEMPURA facilitates more temporally consistent object classification that remains relatively unaddressed in SGG literature. 4) Compared to existing state-of-the-art methods, TEMPURA achieves significant performance gains in terms of mean-Recall@K \cite{VCTree_Tang_2019_CVPR}, highlighting its superiority in generating more unbiased scene graphs.

%% file: long_tail_intro.tex
\begin{figure}[t]
\centering
\begin{subfigure}[b]{0.417\linewidth}
  \centering
  \includegraphics[width=1\textwidth]{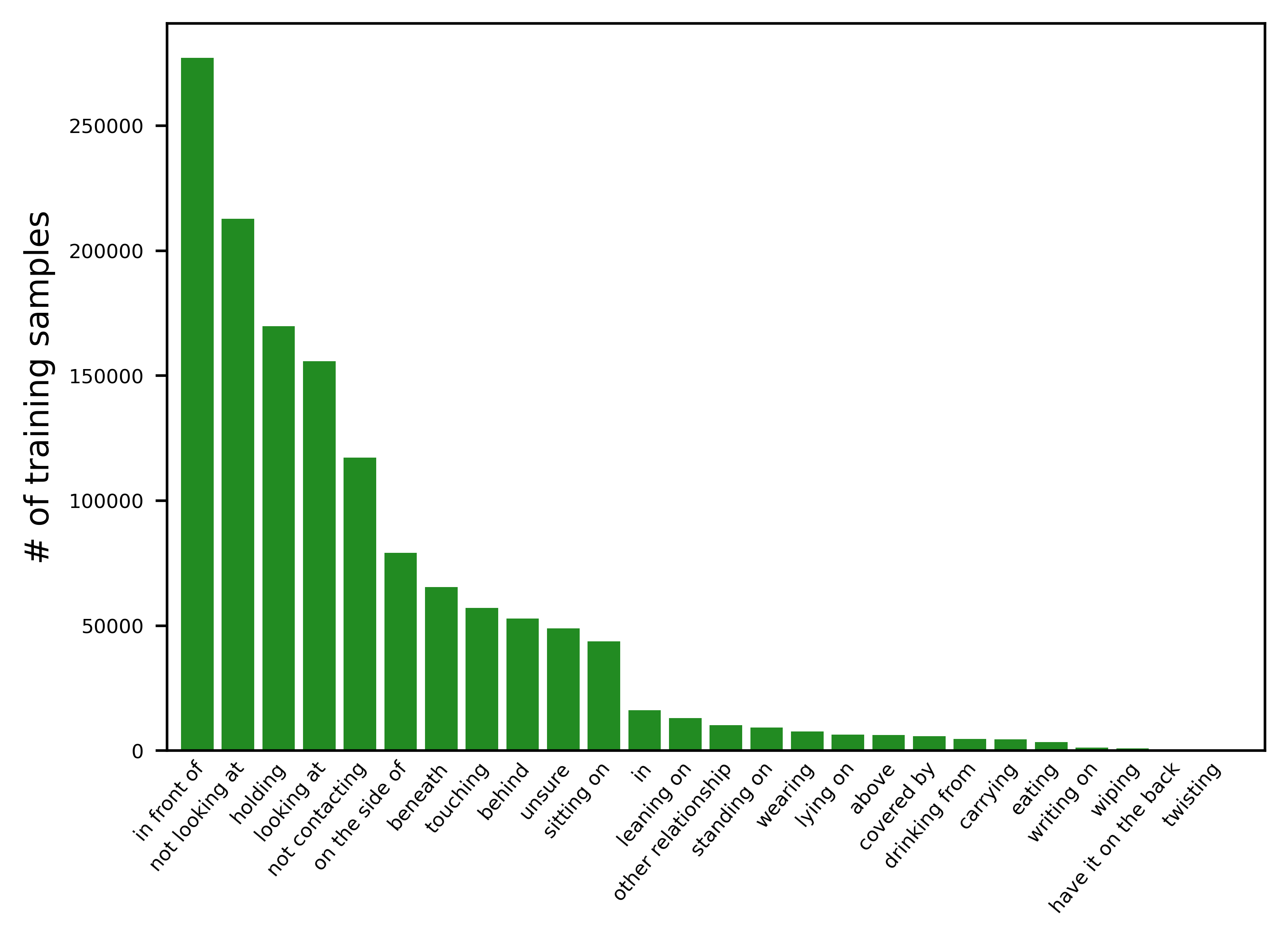} 
  \caption{}
  \label{subfig:long_tail}
\end{subfigure}
\begin{subfigure}[b]{0.42\linewidth}
  \centering
  \includegraphics[width=1\textwidth]{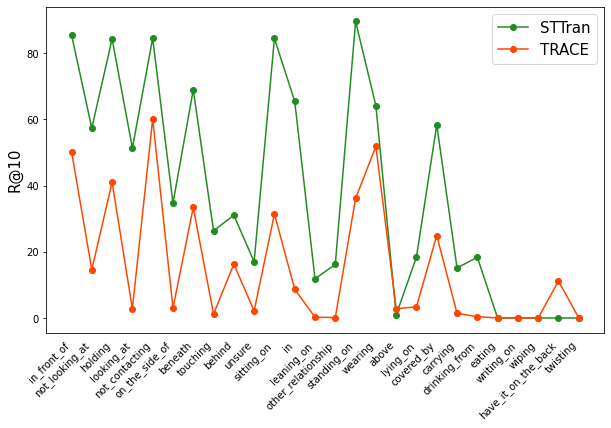} 
  \caption{}
  \label{subfig:sota_behaviour}
\end{subfigure}
\vspace{-0.75em}
\caption{(a) Long-tailed distribution of the predicate classes in Action Genome \cite{Action_genome_Ji_2020_CVPR}. (b) Visual relationship or predicate classification performance of two state-of-the-art dynamic SGG methods, namely STTran \cite{STTran_2021} and TRACE \cite{teng2021target}, falls off significantly for the tail classes.}
\label{fig:long_tail_bias}
\vspace{-1.5em}
\end{figure}

%% file: annotation_characteristics.tex
\begin{figure}[t]
\centering
\begin{subfigure}[b]{0.45\linewidth}
  \centering
  \includegraphics[width=1\textwidth]{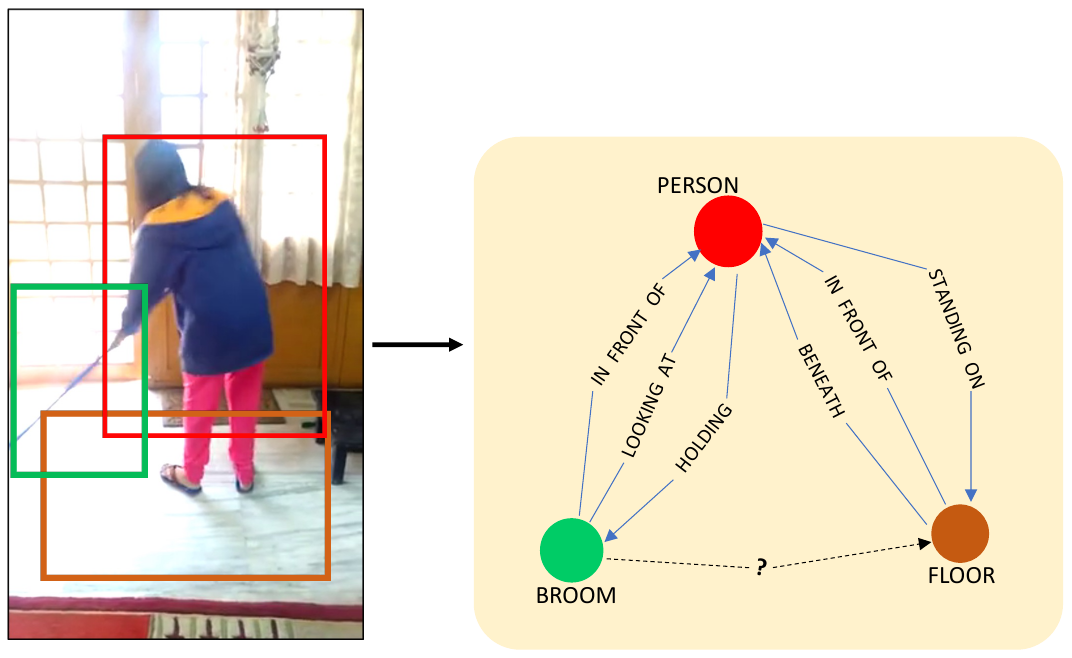} 
  \caption{\centering Incomplete annotations and multiple correct predicates.}
\end{subfigure}
\begin{subfigure}[b]{0.4\linewidth}
  \centering
  \includegraphics[width=1\textwidth]{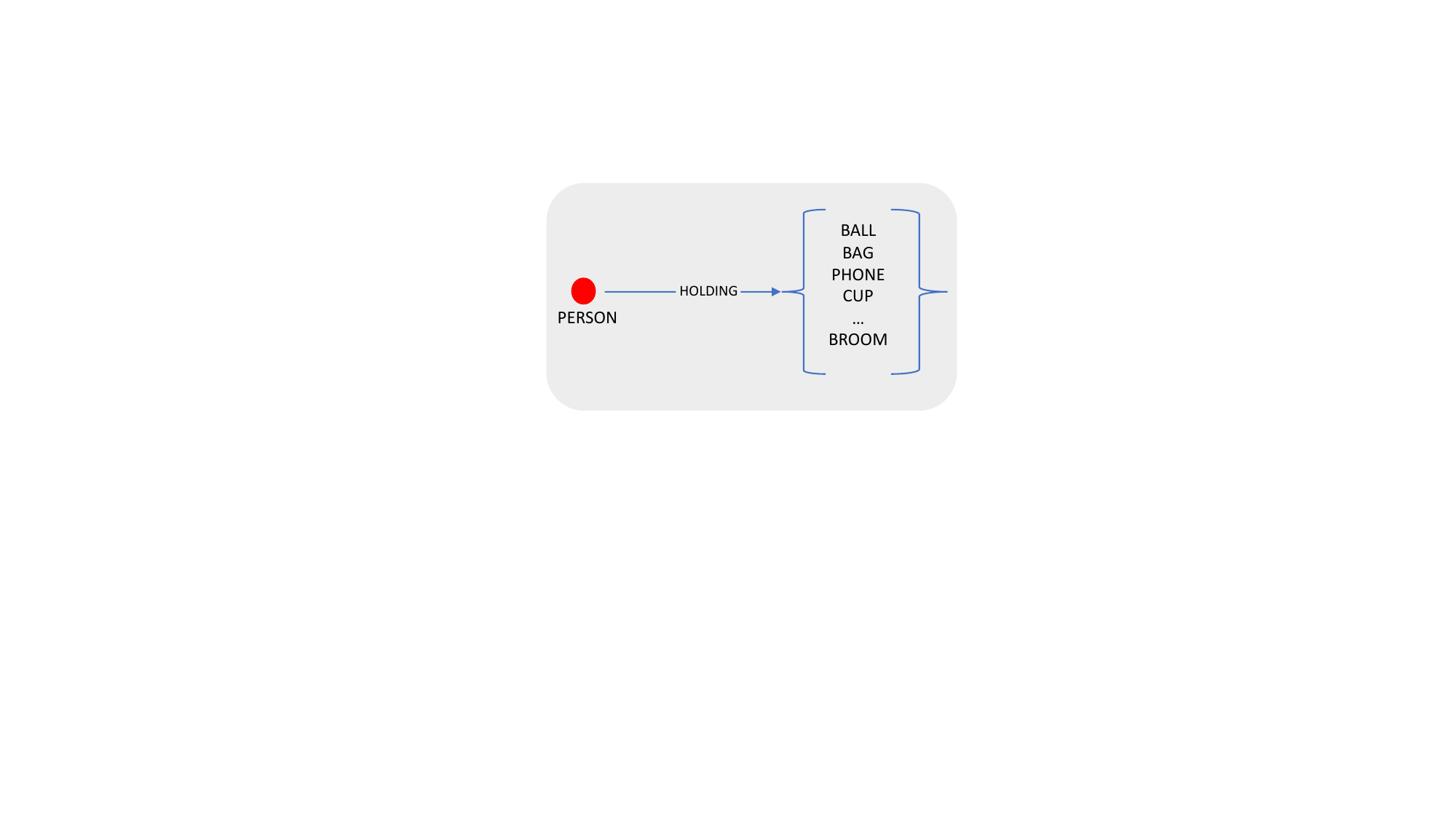}
  \caption{\centering Triplet variability (multiple possible object pairs for the same relationship)}
\end{subfigure}
\vspace{-0.5em}
\caption{
Noisy scene graph annotations in Action Genome \cite{Action_genome_Ji_2020_CVPR} increase the uncertainty of predicted scene graphs. }
\label{fig:noisy_anno}
\vspace{-1.75em}
\end{figure}

%% file: inconsitent_object_clf.tex
\begin{figure}[t!]
\centering
\includegraphics[width=0.28\textwidth]{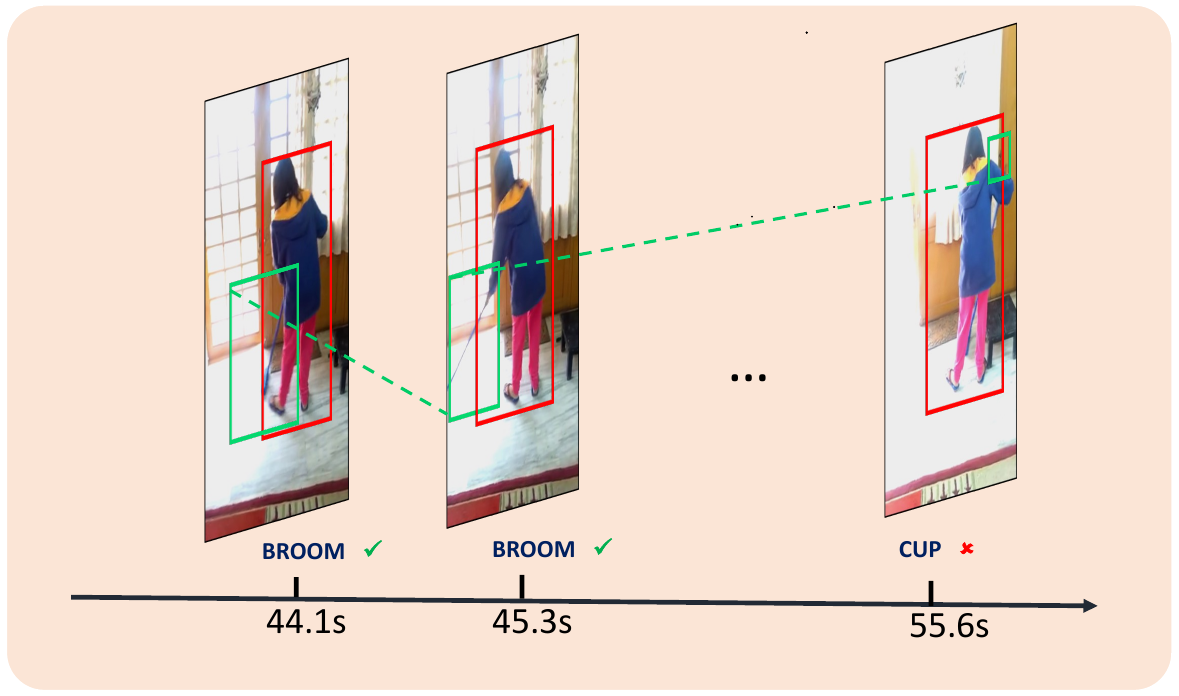} 
\vspace{-0.7em}
  \caption{\small 
  Occlusion and motion blur caused by moving objects in videos renders off-the-self object detectors such as FasterRCNN \cite{ren_frcnn}
  ineffective in producing consistent object classification.}
  \label{fig:inconsistent_obj_clf}
  \vspace{-1.0em}
\end{figure}

%% file: 2_related_works.tex
\section{Related Work}

\noindent\textbf{Image Scene Graph Generation.} SGG from images aims to obtain a graph-structured summarization of a scene where the nodes are objects, and the edges describe their interaction or relationships (formally called predicates). Since the introduction of the image SGG benchmark Visual Genome (VG) \cite{krishnavisualgenome}, research on SGG from \emph{single images} has evolved significantly, with earlier works addressing image SGG utilizing several ways to aggregate spatial context \cite{neural_motifs_2017,Li2017SceneGG,zhang2019vrd,lu2016visual,lin2020gps} and latest ones \cite{tang2020unbiased,wen2020unbiased,VCTree_Tang_2019_CVPR,li2021bipartite,desai2021learning,li2022ppdl,li2022sgtr,li2022rethinking} addressing fundamental problems such as preventing biased scene graphs caused by long-tailed predicate distribution and noisy annotations in image SGG dataset~\cite{krishnavisualgenome}.

\par
\noindent
\textbf{Dynamic Scene Graph Generation.}
Dynamic Scene Graph Generation aims at learning the \emph{spatio-temporal dependencies} of visual relationships between different object pairs over all the frames in a video \cite{Action_genome_Ji_2020_CVPR}. Similar to SGG from images, long-tailed bias and noisy annotations pose a significant challenge to dynamic SGG, further compounded by the temporal fluctuations of predictions. In recent years, a handful of works have attempted to address dynamic SGG \cite{STTran_2021,ji_HORT,teng2021target,TPI_MM_22,APT_Li_2022_CVPR,ISG_arxiv22} and benchmarked their methods on Action Genome (AG) \cite{Action_genome_Ji_2020_CVPR} dataset. While methods like TRACE \cite{teng2021target} introduced temporal context from pretrained 3D models \cite{I3D}, the majority resorted to using the superior sequence processing ability of transformers \cite{vaswani2017attention,videoBERT_Sun_2019_ICCV,arnab2021vivit,nawhal2021activity}  for spatio-temporal reasoning of visual relations. However, despite their success, the performance gains of these methods are mostly realized for the high-frequency relationships and they fail to address the long-tail bias -- the focus of this paper.
 

\noindent\textbf{Mixture Density Networks.} Mixture density networks have been successful in modeling predictive uncertainty and attenuation of noise in many deep-learning tasks. They have been used in many tasks that involve noisy data such as reinforcement learning \cite{choi2018uncertainty}, active learning \cite{choi2021active}, semantic segmentation \cite{Kendall_uncertainty} and even in compensating for data imbalance in image recognition, \cite{imbalanced_uncertainty}. This work is the first to apply such concepts to dynamic SGG.

\noindent\textbf{Memory guided low shot and long-tailed learning.} Memory-guided training strategies \cite{sukhbaatar2015end,mem-aug-meta} have become 
successful 
in addressing learning with data scarcity such as few-shot learning \cite{snell2017prototypical,gidaris2018dynamic,kaiser2017learning} and long tail recognition \cite{parisot2022long,zhu2020inflated}. They enable the learning of generalizable representations by transferring knowledge from data-rich to data-poor classes. We exploit these principles in this paper for learning more unbiased representations of visual relationships in videos.

%% file: 3_method.tex
\input{framework.tex}
\section{Method}
\subsection{Problem Statement}
The goal of dynamic SGG is to describe a structured representation $G_t = \{S_t,R_t,O_t \}$ of each frame $I_t$ in a video $\mathcal{V} = \{I_1,I_2,...,I_T \} $. Here $S_t = \{s_{1}^{t},s_{2}^{t},...,s_{N(t)}^{t} \}$ and $O_t = \{o_{1}^{t},o_{2}^{t},...,o_{N(t)}^{t} \}$ map to the same set of $N(t)$ detected objects in the $t^{th}$ frame. They are combinatorially arranged as subject-object pairs $(s_{j}^{t},r_{k}^{t},o_{i}^{t})$ with $R_t = \{r_{1}^{t},r_{2}^{t},...,r_{K(t)}^{t} \}$ being the set of $K(t)$ predicates describing  the visual relationships between all subject-object pairs in the $t^{th}$ frame.  Formally  each \scalebox{0.95}{ \textless $subject-predicate-object$ \textgreater} or $(s_{j}^{t},r_{k}^{t},o_{i}^{t})$ is called a  triplet. The set of object and predicate classes are referred to as $\mathcal{Y}_{o} = \{y_{o_1},y_{o_2},...,y_{o_{\mathcal{C}_{o}}} \} $ and $\mathcal{Y}_{r} = \{y_{r_1},y_{r_2},...,y_{r_{\mathcal{C}_{r}}} \} $ respectively.

\subsection{Overview}
To generate more unbiased scene graphs from videos, it is necessary to address the challenges highlighted in Fig   \ref{fig:long_tail_bias}, \ref{fig:noisy_anno} and \ref{fig:inconsistent_obj_clf}. To this end, we propose \textbf{TEMPURA} for unbiased dynamic SGG. As shown in Fig \ref{fig:framework}, TEMPURA works with a predicate embedding generator (PEG) that can be obtained from any existing dynamic SGG model \cite{teng2021target,STTran_2021}. Since transformer-based models have shown to be better learners of spatio-temporal dynamics, we model our PEG as the spatio-temporal transformer of \cite{STTran_2021} which is built on top of the vanilla transformer architecture of \cite{vaswani2017attention}. The object sequence processing unit (OSPU) facilitates temporally consistent object classification. The memory diffusion unit (MDU) and the Gaussian Mixture Model (GMM) head address the long-tail bias and overall noise in video SGG data, respectively.  In the subsequent sections, we describe these units in more detail, along with the training and testing mechanism of TEMPURA.

\subsection{Object Detection and Temporal Consistency}
We first describe how we enforce more consistent object classification across the entire video. Using an off-the-self object detector, we obtain  the set of objects $O_t = \{ o_{i}^{t}\}_{i=1}^{N(t)}$ in each frame, where $o_{i}^{t} = \{b_{i}^{t},\bm{v}_{i}^{t},c_{o_i}^{t}\}$ with $b_{i}^{t} \in \mathbb{R}^{4}$ being the bounding box, $\bm{v}_{i}^{t}  \in \mathbb{R}^{2048} $ the RoIAligned \cite{he2017mask} proposal feature of $o_{i}^{t}$ and $c_{o_i}^{t}$ is its predicted class. Existing methods \cite{STTran_2021,ji_HORT,TPI_MM_22,APT_Li_2022_CVPR} either directly use $c_{o_i}^{t} $ as the object classification or pass  $\bm{v}_{i}^{t}$ through a single/multi-layered feed-forward network (FFN) to classify $o_i$. However, object detectors trained on static images fail to compensate for dynamic nuances and temporal fluctuations in videos, making them prone to misclassify the same object in different frames. Some works address this by incorporating object tracking algorithms \cite{teng2021target}, but we incorporate a simple but effective learning-based strategy. 

We introduce an Object Sequence Processing Unit (OSPU) which utilizes a transformer encoder \cite{vaswani2017attention} referred to as sequence encoder or $SeqEnc$ (Fig \ref{fig:framework}), to process a set of sequences, $\mathcal{T}_{\mathcal{V}}$, which is constructed as follows, 
\vspace{-0.25em}
\begin{equation}
\centering
\small
\mathcal{T}_{\mathcal{V}} = \{\mathcal{T}_{t_1k_1}^{1},\mathcal{T}_{t_2k_2}^{2},...,\mathcal{T}_{t_{\hat{\mathcal{C}}_o}k_{\hat{\mathcal{C}}_o}}^{\hat{\mathcal{C}}_o}  \}; \ \ 
\mathcal{T}_{t_jk_j}^{j}= \{\bm{v}_{i}^{t},\bm{v}_{i}^{t+1},...,\bm{v}_{i}^{t+k} \},
\label{eq:sequence}
\vspace{-0.25em}
\end{equation}
where each entry of $\mathcal{T}_{t_jk_j}^{j}$ has the same detected class $c_{o_j}$, $ 1 \leq t_j,k_j \leq T $ and \scalebox{0.9}{$\hat{\mathcal{C}}_o \leq \mathcal{C}_o$} refers to all detected object classes in the video $\mathcal{V}$. Zero-padding is used to turn $\mathcal{T}_{\mathcal{V}}$ into a functioning tensor.
$SeqEnc$ utilizes the multi-head self-attention to learn the long-term temporal dependencies in each \scalebox{0.9}{$\mathcal{T}_{t_jk_j}^{j}$}. For any input $\bm{X}$, a single attention head, $\mathbb{A}$, is defined as follows:
\vspace{-0.5em}
\begin{equation}
    \mathbb{A}(\bm{Q},\bm{K},\bm{V}) = Softmax(\frac{\bm{Q}\bm{K}^{T}}{\sqrt{D_k}})\bm{V} ,
\vspace{-0.15em}
\end{equation}
where $D_k$ is the dimension of $\bm{K}$, and $\bm{Q},\bm{K},\bm{V}$ are the query, key and value vectors which for self-attention is $\bm{Q} = \bm{K} = \bm{V} = \bm{X}$. The multi-head attention, $\mathbb{MA}$, is shown below,
\vspace{-0.3em}
\begin{equation}
\centering
\begin{aligned}
   & \mathbb{MA}(\bm{X}) = Concat(a_1,a_2,..a_H)W_H, \\
   & a_i = \mathbb{A}(\bm{X}W_{Q_i},\bm{X}W_{K_i},\bm{X}W_{V_i}),
\end{aligned}
\label{eq:multihead_1}
\end{equation}
where \scalebox{0.9}{$W_{Q_i} \in \mathbb{R}^{D \times D_{Q_i}}$}, \scalebox{0.9}{$W_{K_i} \in \mathbb{R}^{D \times D_{K_i}}$}, \scalebox{0.9}{$W_{V_i}\in\mathbb{R}^{D \times D_{V_i}}$} and \scalebox{0.9}{$W_H \in \mathbb{R}^{HD_v \times D}$} are learnable weight matrices. As shown in Fig \ref{fig:framework}, we follow the classical design of \cite{vaswani2017attention} for  $SeqEnc$, whereby a residual connection is used to add $\bm{V}$ with $\mathbb{MA}(\bm{X})$ followed by normalization \cite{Layer_norm}, and subsequent passing through an FFN. The output of an $n$ layered sequence encoder is as follows,
\vspace{-0.5em}
\begin{equation}
\centering
\bm{X}_{out}^{(n)} = SeqEnc(\bm{X}_{out}^{(n-1)}); \ \ \bm{X}_{out}^{(0)} = \hat{\mathcal{T}}_{\mathcal{V}},
\label{eq:encoder}
\vspace{-0.25em}
\end{equation}
where \scalebox{0.9}{$\hat{\mathcal{T}}_{\mathcal{V}} = \mathcal{T}_{\mathcal{V}}+\bm{E}_o^T$} with $\bm{E}_o^T$ being fixed positional encodings \cite{vaswani2017attention} for injecting each object's temporal position. The final object logits, $\mathcal{\hat{Y}}_o = \{\hat{y}_{o_i} \}_{i=1}^{\mathcal{C}_o}$, are obtained by passing $\bm{X}_{out}^{(n)}$ through a $2$-layer FFN. The corresponding object classification loss, $\mathcal{L}_o$, is modeled as the cross-entropy between $\mathcal{\hat{Y}}_o$ and $\mathcal{Y}_o$.

To enhance the $SeqEnc$'s capability of enforcing temporal consistency, we add a supervised contrastive loss \cite{hadsell2006dimensionality} over its output embeddings, as shown below, 
\vspace{-0.4em}
\begin{equation}
\resizebox{0.9\linewidth}{!}{
    $\mathcal{L}_{intra} = \sum\limits_{i}\sum\limits_{j} ||\bm{\hat{x}}_{o_i} - \bm{\hat{x}}_{o_j}^{+} ||_2^{2} + \sum\limits_{k} max(0,1 - ||\bm{\hat{x}}_{o_i} - \bm{\hat{x}}_{o_k}^{-} ||_2^{2})$},
\vspace{-0.25em}
\end{equation}
where \scalebox{0.9}{$\bm{\hat{x}}_{o_i} \in \bm{X}_{out}^{(n)}$}. $\mathcal{L}_{intra}$ enforces intra-video temporal consistency by pulling closer the embeddings of positive pairs sharing the same ground-truth class and pushing apart the embeddings of negative pairs with different ground-truth class.

\subsection{Predicate Embedding Generator}
A predicate embedding generator (PEG) assimilates the information of each subject-object pair to generate an embedding that summarizes the relationship(s) between them. For dynamic SGG, the PEG must learn the temporal as well as the spatial context of the relationship between each pair. In our setup, we model the PEG as the Spatio-Temporal transformer of \cite{STTran_2021}. For each pair $(i,j)$, we construct the input to the PEG as shown below,
\begin{equation}
\small
    \bm{r}_{k}^{t} = Concat(f_v(\bm{v}_{i}^{t}),f_v(\bm{v}_{j}^{t}),f_u(\bm{u}_{ij}^{t} + f_{box}(b_{i}^{t},b_{j}^{t})),\bm{s}_i^{t},\bm{s}_j^{t}),
\label{eq:rel_input}
\end{equation}
where $\bm{v}_{i}^{t}$ and $\bm{v}_{j}^{t}$ are the subject and object proposal features, $\bm{u}_{ij}^{t} \in \mathbb{R}^{256 \times 7 \times 7}$ is the feature map of the union box computed by RoIAlign \cite{he2017mask}, $\bm{s}_i^{t}$,$\bm{s}_j^{t} \in \mathbb{R}^{200}$ are the semantic glove embeddings \cite{pennington2014glove} of the subject and object class determined from $\mathcal{\hat{Y}}_o$, $f_v$ and $f_u$ are FFN based non-linear projections, $f_{box}$ is the bounding box to feature map projection of \cite{neural_motifs_2017}. The set of $t^{th}$ frame input representations are $\bm{R}_t=\{\bm{r}_t^{j}\}_{j=1}^{K(t)} \in \mathbb{R}^{K(t) \times 1936} $. As shown in Fig \ref{fig:framework} the PEG consists of a spatial encoder, $SpaEnc$, and a temporal decoder, $TempDec$, where the former learns the spatial context of the visual relations and the latter learns their temporal dependencies. Therefore for an $n$ layered spatial encoder, its output $\bm{R}_{spa}^{t}$ is computed as follows,
\vspace{-0.25em}
\begin{equation}
\centering
\bm{Z}_{spa,t}^{(n)} = SpaEnc(\bm{Z}_{spa,t}^{(n-1)}); \  \bm{Z}_{spa,t}^{(0)} = \bm{R}_t \ ,
\label{eq:spa_encoder}
\vspace{-0.25em}
\end{equation}
where $\bm{R}_{spa}^{t} = \bm{Z}_{spa,t}^{(n)}$. The formulation of $SpaEnc$ is the same as $SeqEnc$ (Eq \ref{eq:encoder}). To learn the temporal dependencies of the relationships, the decoder input is constructed as a sequence over a non-overlapping sliding window whereby,
\begin{equation}
\centering
    \bm{Z}_{tem} = \{\bm{R}_{spa}^{t},...,\bm{R}_{spa}^{t+\eta-1} \}, \ \  t \in [1,T-\eta +1],
\end{equation}
where $\eta \leq T$ is the sliding window and $T$ is the length of the video. As shown in Fig \ref{fig:framework}, the inputs to $TempDec$'s $\mathbb{MA}$ are, $\bm{Q}=\bm{K}=\bm{Z}_{tem}+\bm{E}_r^{\eta}$ and $\bm{V}=\bm{Z}_{tem}$ where $\bm{E}_r^{\eta} = \{\bm{e}_r^1,\bm{e}_r^2,...,\bm{e}_r^{\eta}  \}$ are learnable temporal encodings \cite{STTran_2021} injecting the temporal position of each predicate. The final output $\mathcal{R}_{tem}$ of an $n$ layered temporal decoder is,
\vspace{-0.4em}
\begin{equation}
    \bm{Z}_{tem}^{(n)} = TempDec(\bm{Z}_{tem}^{(n-1)});  \  \bm{Z}_{tem}^{(0)}=\bm{Z}_{tem} \ ,
\end{equation}
Therefore, the final set of predicate embeddings generated by the PEG  is $\mathcal{R}_{tem} = \bm{Z}_{tem}^{(n)} = \{ \bm{R}_{tem}^{t} \}_{t=1}^{T-\eta +1}$ with $ \bm{R}_{tem}^{t} = \{\bm{r}_{tem}^{j} \}_{j=1}^{K(t)} \in \mathbb{R}^{K(t) \times 1936}$.

\subsection{\scalebox{1}{ Memory guided Debiasing} }\label{sec:MDU} 

Due to the long-tailed bias in SGG datasets, the direct PEG embeddings, $\mathcal{R}_{tem}$, are biased against the rare predicate classes, necessitating the need to debias them. We accomplish this via  a memory-guided training strategy, whereby for any given relationship embedding, $\bm{r}_{tem}^{j} \in \mathcal{R}_{tem}$, a Memory Diffusion Unit (MDU) first retrieves relevant information from a predicate class centric memory bank $\bm{\Omega}_{R}$ and uses it to enrich $\bm{r}_{tem}^{j}$ which results in a more balanced embedding $\bm{\hat{r}}_{tem}^{j}$. The memory bank $\bm{\Omega}_{R} = \{\bm{\omega}_p\}_{p=1}^{\mathcal{C}_r}$ is composed of a set of memory prototypes each of which is an abstraction of a predicate class and is computed as a function of their corresponding PEG embeddings. In our setup, the prototype is defined as a class-specific centroid, whereby, \scalebox{0.82}{$\bm{\omega}_p = \frac{1}{N_{y_{r_p}}}\sum\limits_{j=1}^{N_{y_{r_p}}}\bm{r}_{tem}^{j} \ \ \forall \ p \in \mathcal{Y}_r$}, with $N_{y_{r_p}}$ being the total number of subject-object pairs mapped to the predicate class $y_{r_p}$, in the \emph{entire training set}.

\noindent\textbf{Progressive Memory Computation.} $\bm{\Omega}_{R}$ is computed in a progressive manner whereby the model's last state is used to compute memory for the current state, i.e., the memory of epoch $\alpha$ is computed using the model weights of epoch $\alpha-1$. This enables $\bm{\Omega}_{R}$ to become more refined with every epoch. Since no memory is available for the first epoch, the MDU remains inactive for this state, and $\bm{\hat{r}}_{tem}^{j} = \bm{r}_{tem}^{j}$. 


\input{mdu.tex}

\noindent\textbf{Memory Diffusion Unit.} As shown in Fig \ref{fig:MDU} for a given query the MDU uses the attention operator \cite{vaswani2017attention} to retrieve relevant information from $\bm{\Omega}_R$ as a diffused memory feature $\bm{r}_{mem}^j$ i.e.,
\vspace{-0.5em}
\begin{equation}
\bm{r}_{mem}^j = \mathbb{A}(\bm{Q}W_{Q}^{mem},\bm{K}W_{K}^{mem},\bm{V}W_{V}^{mem}),
\label{eq:diffused_mem_feature}
\vspace{-0.5em}
\end{equation}
where, $\bm{Q}=\bm{r}_{tem}^{j}$ and $ \bm{K}=\bm{V}=\bm{\Omega_r}$ and $W_{Q}^{mem}, W_{K}^{mem} \ \ \& W_{V}^{mem}  \in \mathbb{R}^{1936 \times 1936}$ are learnable weight matrices. Since each subject-object pair has multiple predicates mapped to it, many visual relations share similar characteristics, which means their corresponding memory prototypes $\bm{\omega}_p$ share multiple predicate embeddings. Therefore the attention operation of Eq \ref{eq:diffused_mem_feature} facilitates knowledge transfer from data-rich to data-poor classes utilizing the memory bank, whereby $\bm{r}_{mem}^j$ hallucinates compensatory information about the data-poor classes otherwise missing in $\bm{r}_{tem}^{j}$. This information is diffused back to $\bm{r}_{tem}^{j}$ to obtain the balanced embedding $\bm{\hat{r}}_{tem}^{j}$ as shown below,
\vspace{-0.3em}
\begin{equation}
    \bm{\hat{r}}_{tem}^{j} = \lambda\bm{r}_{tem}^{j} + (1-\lambda)\bm{r}_{mem}^j \ ,
    \label{eq:mem_diffusion}
    \vspace{-0.25em}
\end{equation}
where $0 < \lambda \leq 1$. As shown in Fig \ref{fig:framework}, the MDU is used during the training phase only since it does not function as a network module to forward pass through but rather as a meta-learning inspired \cite{parisot2022long,snell2017prototypical,zhu2020inflated} structural meta-regularizer. Since $\bm{\Omega}_R$ is computed directly from the PEG embeddings, backpropagating over the MDU refines the computed memory prototypes, in turn enabling better information diffusion and inherently teaching the PEG how to generate more balanced embeddings that do not underfit to the data-poor relationships. $\lambda$ over here acts as a gradient scaling factor, which during backpropagation asymmetrically scales the gradients associated with $\bm{r}_{tem}^{j}$ and $\bm{r}_{mem}^j$ in the residual operation of Eq \ref{eq:mem_diffusion}. Since the initial PEG embeddings are heavily biased towards the data-rich classes, if $\lambda$ is too high, the compensating effect of the diffused memory feature is drastically reduced. On the other hand, if $\lambda$ is too low, excessive knowledge gets transferred from the data-rich to the data-poor classes resulting in poor performance on the former.

\subsection{\scalebox{1}{Uncertainty Attenuated Predicate Classification }}
To address the noisy annotations in SGG data, we model the predicate classification head as a $\mathcal{K}$ component Gaussian Mixture Model (GMM) \cite{Kendall_uncertainty}. Given a sample embedding $\textbf{z}_i$ the mean, variance and mixture weights for the $p^{th}$ predicate class are estimated as follows:
\vspace{-0.5em}
\begin{equation}
   \small
      \mu_{i,p}^{k} = f_{\mu}^{k}(\mathbf{z}_i),   \  \Sigma_{i,p}^{k} = \sigma(f_{\Sigma}^{k}(\mathbf{z}_i)), \ \pi_{i,p}^{k} = \frac{e^{f_{\pi}^{k}(\mathbf{z}_i)}}{\sum\limits_{k=1}^{\mathcal{K}}e^{f_{\pi}^{k}(\mathbf{z}_i)}},
\end{equation}
where $f_{\mu}^{k}, f_{\Sigma}^{k}, f_{\pi}^{k} $ are FFN based projection functions and $\sigma$ is the sigmoid non-linearity which ensures $\Sigma_{i,p}^{k} \geq 0$. The class-specific aleatoric and epistemic uncertainty, for the sample $\mathbf{z}_i$ are computed as follows: 
\vspace{-0.25em}
\begin{equation}
\scriptstyle
U_{al}^{p}(\textbf{z}_i) = \sum\limits_{k=1}^{\mathcal{K}} \pi_{i,p}^{k}\Sigma_{i,p}^k \ \ ; \ \    
U_{ep}^{p}(\textbf{z}_i) = \sum\limits_{k=1}^{\mathcal{K}} \pi_{i,p}^{k}||\mu_{i,p}^{k} - \sum\limits_{j=1}^{\mathcal{K}}\pi_{i,p}^{j}\mu_{i,p}^{j}||_2^2 \ ,
\vspace{-0.25em}
\end{equation}
 Therefore, by using a GMM head, we are modeling the inherent uncertainty associated with the data from a Bayesian perspective \cite{Kendall_uncertainty,choi2018uncertainty,choi2021active}. During training $\textbf{z}_i = \bm{\hat{r}}_{tem}^{i}$ and the probability distribution for the $p^{th}$ predicate is given as,
 \vspace{-0.5em}
\begin{equation}
    \hat{y}_{r_p}^{i} = \sum\limits_{k=1}^{\mathcal{K}}\pi_{i,p}^{k} \mathcal{N}(\mu_{i,p}^{k},\Sigma_{i,p}^{k}),
\vspace{-0.25em}
\end{equation}
where $\mathcal{N}$ is the Gaussian distribution. Since the sampling $\mathcal{N}(\mu_{p}^{k},\Sigma_{p}^{k})$ is non-differentiable we use the re-parameterization  trick of \cite{kingma2013auto} to compute $\hat{y}_{r_p}^{i}$ as shown below:
\vspace{-0.5em}
\begin{equation}
\small
\hat{y}_{r_p}^{i} = \sum\limits_{k=1}^{\mathcal{K}}\pi_{i,p}^{k}\sigma(\hat{c}_{p,k}^{i}); \ \ \ \hat{c}_{p,k}^{i} = \mu_{i,p}^{k}+\varepsilon \sqrt{\Sigma_{i,p}^{k}} \ ,
\end{equation}
where $\varepsilon \sim \mathcal{N}(0,1) $ and is of the same size as $\Sigma_p^k$. The overall set of predicate logits is $\hat{\mathcal{Y}}_r = \{ \hat{y}_{r_p}\}_{p=1}^{\mathcal{C}_r}$. The predicate classification loss $\mathcal{L}_p$ is modeled as the GMM sigmoidal cross entropy \cite{choi2021active} as shown below, 
\begin{equation}
    \mathcal{L}_p = -\sum\limits_{i=1}^{N_{r,p}} \sum\limits_{p=1}^{\mathcal{C}_r} y_{r_p}^{i} \log \sum\limits_{k=1}^{\mathcal{K}}\pi_{p}^{k}\sigma(\hat{c}_{p,k}^{i}),
    \label{eq:predicate_classification}
\end{equation}
where $y_{r_p}^{i}$ is the ground-truth predicate class mapped to $\bm{z}_i$. 
By incorporating the modeled aleatoric uncertainty of $\mathbf{z}_i$ ($\Sigma_{i,p}^k$) in $\mathcal{L}_p$, we essentially utilize it as an \emph{attenuation} factor, which penalizes the model if $\Sigma_{i,p}^k$ is large. This principle is called \emph{learned loss attenuation} \cite{Kendall_uncertainty,choi2018uncertainty}, and it discourages the model from predicting high uncertainty thereby attenuating the effects of uncertain samples due to inherent annotation noise in the data. 

\subsection{Training and Testing}
\noindent\textbf{Training.} As explained in section 3.5, memory computation and utilization of MDU is activated from the second epoch, and so for the first epoch, $\bm{\hat{r}}_{tem}^i = \bm{r}_{tem}^i$. 
The OSPU and the GMM head obviously start firing from the first epoch itself. The entire framework is trained end-to-end by minimizing the following loss,
\vspace{-0.25em}
\begin{equation}
    \mathcal{L}_{total} = \mathcal{L}_p + \mathcal{L}_o + \mathcal{L}_{intra} \ ,
\vspace{-0.25em}
\end{equation}

\noindent\textbf{Testing.} The forward pass during testing is highlighted in Fig \ref{fig:framework}. After training, the MDU has served its purpose of teaching the PEG to generate more unbiased embeddings, and therefore during inference, $\bm{r}_{tem}^i$ is directly passed to the GMM head to obtain the predicate confidence scores, $\hat{y}_{r_p}^i$, which during testing are  computed as follows,
\vspace{-0.5em}
\begin{equation}
    \hat{y}_{r_p}^i = \sum\limits_{k=1}^{\mathcal{K}}\pi_{i,p}^k \sigma(\mu_{i,p}^k),
\vspace{-1.25em}
\end{equation}.

%% file: framework.tex
\begin{figure*}
\begin{center}
\includegraphics[width=0.84\linewidth]{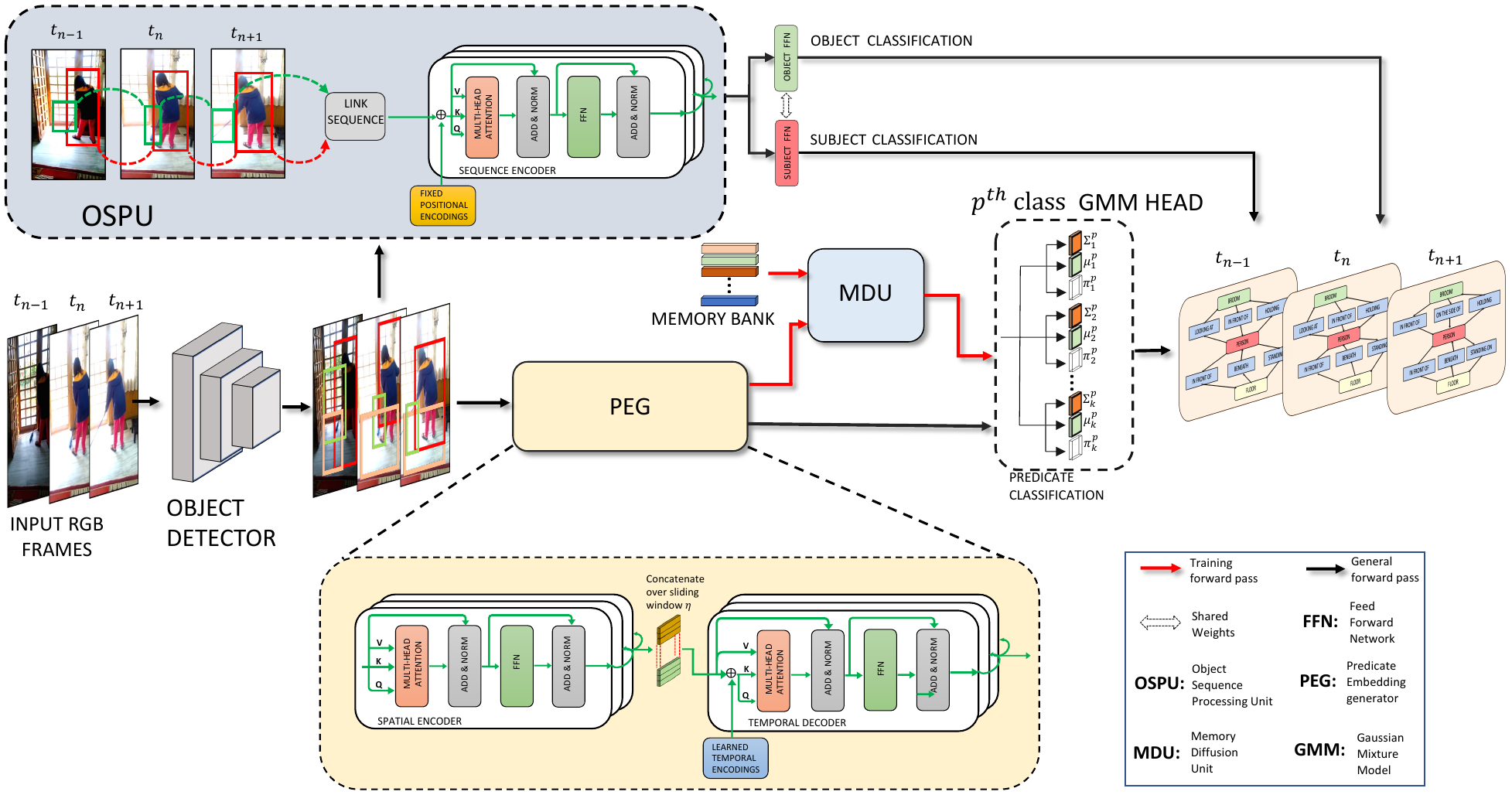}
\end{center}
\vspace{-1.1em}
   \caption{ {\bf Framework of TEMPURA.} 
   The object detector generates initial object proposals for each RGB frame in a video. The proposals are then passed to the OSPU, where they are first linked into sequences based on the object detector's confidence scores. These sequences are processed with a transformer encoder to generate temporally consistent object embeddings for improved  object classification. The proposals and semantic information of each subject-object pair are passed to the PEG to generate a spatio-temporal representation of their relationships. Modeled as a spatio-temporal transformer \cite{STTran_2021}, the PEG's encoder learns the spatial context of the relationships and its decoder learns their temporal dependencies. Due to the long-tail nature of the relationship/predicate classes, a Memory Bank in conjunction with the MDU is used during training to debias the PEG, enabling the production of more generalizable predicate embeddings. Finally, a $\mathcal{K}$-component GMM head classifies the PEG embeddings and models the uncertainty associated with each predicate class for a given subject-object pair.}
\label{fig:framework}
\vspace{-1.35em}
\end{figure*}

%% file: mdu.tex
\begin{figure}[t]
\begin{center}
\includegraphics[width=0.8 \linewidth]{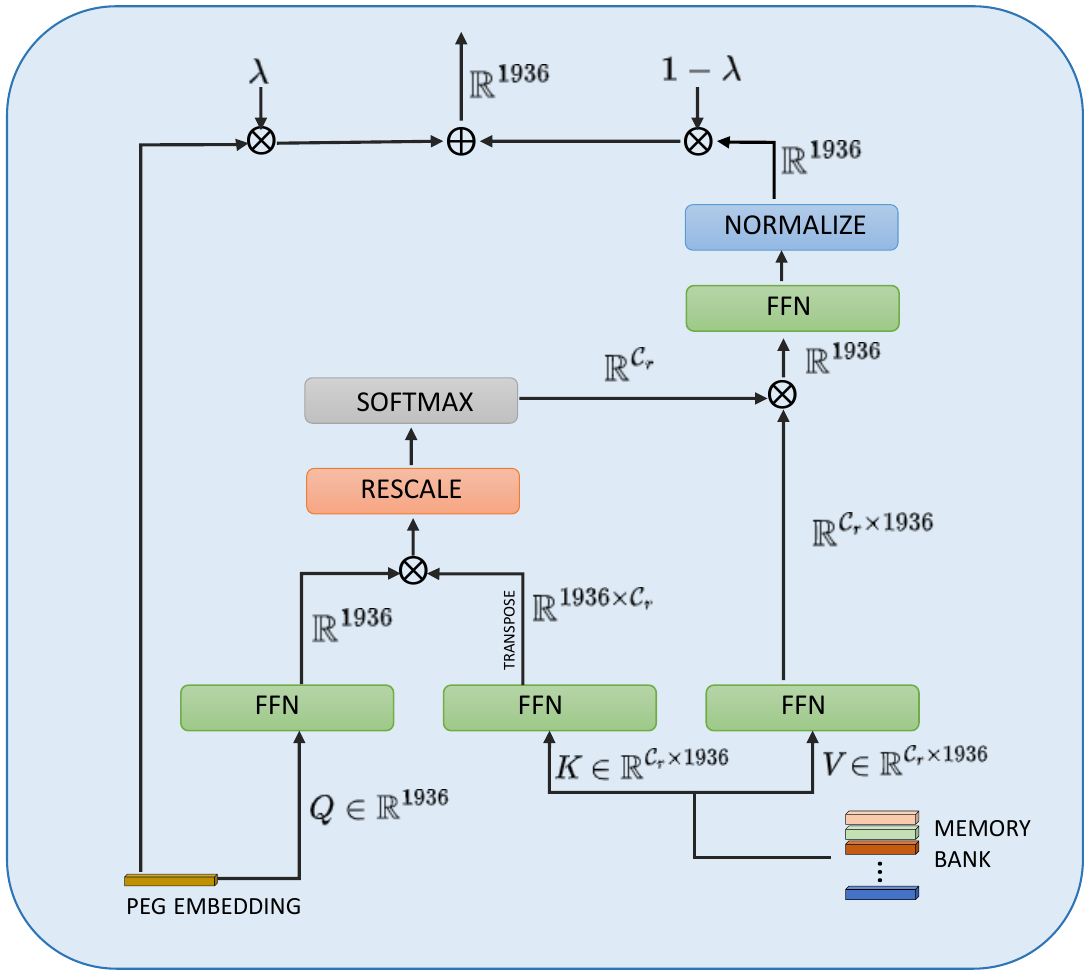}
\end{center}
   \vspace{-1.2em}
   \caption{\textbf{Illustration of the Memory Diffusion Unit (MDU)}. $\otimes$ and $\oplus$ are matrix multiplication and element-wise addition respectively. }
\label{fig:MDU}
\vspace{-1.1em}
\end{figure}

%% file: 4_experiments.tex
\section{Experiments}

\input{sgdet_all_sota.tex}

\input{mR_sota_pcls_sgcls.tex}
\input{R_sota_pcls_sgcls.tex}

\subsection{Dataset and Implementation}

\noindent\textbf{Dataset.} We perform experiments on the Action Genome  (AG) \cite{Action_genome_Ji_2020_CVPR} dataset, which is the largest benchmark dataset for video SGG. It is built on top of Charades \cite{Charades} and has $234,253$ annotated frames with $476,229$ bounding boxes for $35$ object classes (without person), with a total of $1,715,568$ annotated predicate instances for $26$ relationship classes.

\noindent\textbf{Metrics and Evaluation Setup.}
We evaluate the performance of TEMPURA with standard metrics Recall@K (R@K) and mean-Recall@K (mR@K), for $K =[10,20,50] $. As discussed before, R@K tends to be biased towards the most frequent predicate classes \cite{VCTree_Tang_2019_CVPR} whereas mR@K is a more balanced metric enabling evaluation of SGG performance on all the relationship classes \cite{VCTree_Tang_2019_CVPR}. As per standard practice \cite{Action_genome_Ji_2020_CVPR,krishnavisualgenome,STTran_2021,teng2021target}, three SGG tasks are chosen, namely: (1) Predicate classification (\textit{PREDCLS}): Prediction of predicate labels of object pairs, given ground truth labels and bounding boxes of objects; (2) Scene graph classification (\textit{SGCLS}): Joint classification of predicate labels and the ground truth bounding boxes; (3) Scene graph detection (\textit{SGDET}): End-to-end detection of the objects and predicate classification of object pairs. 
Evaluation is conducted under two setups:  \textbf{With Constraint} and \textbf{No constraints}. In the former the generated graphs are restricted to at most one edge, i.e., each subject-object pair is allowed only one predicate and in the latter, the graphs can have multiple edges. We note that mean Recall is averaged over all predicate classes, thus reflective of an SGG model's long-tailed performance as opposed to Recall, which might be biased towards head classes. 

\noindent\textbf{Implementation details.}
Following prior work, \cite{STTran_2021,APT_Li_2022_CVPR}, we choose FasterRCNN \cite{ren_frcnn} with ResNet-101 \cite{Resnet} as the object detector. For the predicate embedding generator, we choose the Spatio-temporal transformer architecture of \cite{STTran_2021}, with the same number of encoder-decoder layers and attention heads. The gradient scaling factor $\lambda$ is set to $0.5$ for \textit{PREDCLS} and \textit{SGDET} and $0.3$ for \textit{SGCLS}. The number of GMM components $\mathcal{K}$ is set to $4$ for \textit{SGCLS} and \textit{SGDET} and $6$ for PREDCLS. The framework is trained end to end for $10$ epochs using the AdamW optimizer \cite{AdamW} and a batch size of $1$. The initial learning rate is set to $10^{-5}$.

\subsection{Comparison to state-of-the-art}

We compare our method with existing dynamic SGG methods such as STTran \cite{STTran_2021}, TRACE \cite{teng2021target}, STTran-TPI \cite{TPI_MM_22}, APT \cite{APT_Li_2022_CVPR},  ISGG \cite{ISG_arxiv22}. We also compare with ReLDN \cite{RelDN} which is a static SGG method. 
Table \ref{tab:sgdet_all} shows the comparative results for \textit{SGDET} in terms of both mR@K and R@K.
Tables \ref{tab:mR_sota_predcls_sgcls} and \ref{tab:R_sota_predcls_sgcls} show the comparative results for \textit{PREDCLS} $+$ \textit{SGCLS} in terms of mR@K and R@K respectively. 
We utilized the official code for several state-of-the-art dynamic SGG methods to obtain respective mR@K values for all three SGG tasks under both \textbf{With Constraint} and \textbf{No Constraints} setup. 
We also relied on email communications with the authors of several papers on the mR values where the source code are not publicly available. 
From Tables \ref{tab:sgdet_all} and \ref{tab:mR_sota_predcls_sgcls}, we observe that TEMPURA significantly outperforms the other methods in mean Recall. Specifically, in comparison to the best baseline, we observe improvements of $\textbf{5.1} \%$ on \textit{PREDCLS}-mR@10, $\textbf{5.7} \%$ on \textit{SGCLS}-mR@10 and $\textbf{1.9} \%$ on \textit{SGDET}-mR@10 under the \textbf{With Constraint} setup. For the \textbf{No Constraints} setup the improvements are even more significant with $\textbf{10.1} \%$ on \textit{PREDCLS}-mR@10, $\textbf{7.6} \%$ on \textit{SGCLS}-mR@10 and $\textbf{3.8} \%$ on \textit{SGDET}-mR@10. This clearly shows that TEMPURA can generate more unbiased scene graphs by better detecting both data-rich and data-poor classes. This is further verified from Fig \ref{fig:per_class_mR} where we compare mR@10 values for the \emph{HEAD}, \emph{BODY} and \emph{TAIL} classes of AG with that of STTran and TRACE. TEMPURA significantly improves performance on the \emph{TAIL} classes without compromising performance on the \emph{HEAD} and \emph{BODY} classes. Similar charts for the \textbf{No Constraints} setup are provided in the supplementary. The comparative per-class performance in Fig \ref{fig:per_class_line} further shows that TEMPURA outperforms both STTran and TRACE for most predicate classes.
Tables \ref{tab:sgdet_all} and \ref{tab:R_sota_predcls_sgcls} show that TEMPURA does not compromise Recall values and achieves comparable or better performance than the existing methods, which made  deliberate efforts to achieve high Recall values without considering their long-tailed performances. Qualitative visualizations are shown in Fig \ref{fig:qualitative}.

\input{mR_per_class.tex}
\input{predcls_per_class_line}

\input{qualitative_results.tex}

\input{sgcls_sgdet_ablations.tex}
\input{K_ablation.tex}

\subsection{Ablation Studies}
We conduct ablation experiments on \textit{SGCLS} and \textit{SGDET} tasks to study the impact of the OSPU, MDU, and GMM head, the combination of which enables TEMPURA to generate more unbiased scene graphs. When all these components are removed, TEMPURA essentially boils down to the baseline STTran architecture \cite{STTran_2021}, where the object proposals and PEG embeddings are mapped to a few layers of FFN for respective classification.

\noindent\textbf{Uncertainty Attenuation and Memory guided Training.} We first study the impact of uncertainty-aware learning and memory-guided debiasing. For the first case, we remove the MDU during training and use only the GMM head. For the second case, we substitute the GMM head with a simple FFN head as the classifier, with the predicate loss $\mathcal{L}_p$ converted to a simple multi-label binary cross entropy. The results of these respective cases are shown in rows $1$ \& $2$ of Table \ref{tab:sgcls_sgdet_ablations}. It can be observed that the resulting models improve mR@K performance over the baseline. This indicates two things: 1) Modeling and attenuation of the predictive uncertainty of an SGG model can effectively address the noise associated with the TAIL classes, preventing it from under-fitting to them \cite{imbalanced_uncertainty}. 2) MDU-guided training enables the PEG to generate embeddings that are more robust and generalizable to all the predicate classes, which performs slightly better than just using uncertainty-aware learning for all three SGG tasks. Combining both these principles gives the best performance, as seen in the final row of both tables.




\noindent\textbf{Temporally Consistent Object Classification.} By comparing rows $3$ and $4$ of Table \ref{tab:sgcls_sgdet_ablations}, we can see that without the OSPU-based enforcement of temporal consistency on object classification, the performance drops significantly, highlighting the fact that object misclassification due to temporal nuances in videos is also a major source of noise in existing SGG frameworks. For the \textit{PREDCLS} task the ground-truth bounding boxes and labels are already provided, so the OSPU has no role, and its weights are frozen during training.

\noindent\textbf{Number of Gaussian components} $\mathcal{K}$. The performance of TEMPURA for different values of $\mathcal{K}$ is shown in Table \ref{tab:K_ablation}. Keeping $\mathcal{K}$ b/w $4$ and $6$ gives the best performance, beyond which the model incurs a heavy memory footprint with diminishing returns. 
More ablation experiments are provided in the supplementary.



%% file: sgdet_all_sota.tex
\begin{table*}[tb!]
\centering
\caption{ Comparative results for SGDET task, on AG \cite{Action_genome_Ji_2020_CVPR}, in terms of mean-Recall@K and Recall@K. Best results are shown in bold. }
\setlength{\tabcolsep}{12pt}
\vspace{-0.5em}
\resizebox{1\linewidth}{!}
{%
\begin{tabular}{ccccccccccccc}
\toprule
  \multirow{2}{*}{Method} & \multicolumn{6}{c}{With Constraint} & \multicolumn{6}{c}{No Constraints}\\
  \cmidrule(lr){2-7} \cmidrule(lr){8-13}   
   & mR@10 & mR@20 & mR@50 &R@10 &R@20 &R@50 & mR@10 & mR@20 & mR@50 &R@10 &R@20 &R@50 
\\
\midrule
\midrule
  RelDN~\cite{zhang2019vrd} & 3.3 & 3.3 & 3.3& 9.1 & 9.1 & 9.1 & 7.5 & 18.8 & 33.7 & 13.6 & 23.0 & 36.6  \\
  HCRD supervised\cite{simple_baseline_bmvc} &- & 8.3 & 9.1 &  - & 27.9  & 30.4 & - & - &  - & - & - &  -   \\
  TRACE \cite{teng2021target}& 8.2 & 8.2 & 8.2& 13.9 &	14.5 &	14.5  &  22.8 &31.3 & 41.8 & 26.5 & 35.6 &	45.3 \\
  
  ISGG \cite{ISG_arxiv22} &  - & 19.7 & 22.9 & - & 29.2 & 35.3  & - & - & - & - & - & -\\ 
  STTran \cite{STTran_2021}&  16.6 & 20.8 & 22.2&   25.2 & 34.1 & 37.0  & 20.9 & 29.7 & 39.2& 24.6 &	36.2 & 48.8  \\
  
  STTran-TPI \cite{TPI_MM_22} & 15.6 & 20.2 & 21.8& 26.2 & 34.6 & 37.4  & - & - & - & - & - & -\\

  APT \cite{APT_Li_2022_CVPR}  & - & - & -& 26.3 & \textbf{36.1}	& \textbf{38.3} &  - & - & - &25.7 &	37.9 &	\textbf{50.1}\\
  
  TEMPURA &  \textbf{ 18.5} &\textbf{ 22.6} & \textbf{23.7} & \textbf{28.1} &	33.4 &	34.9  &  \textbf{24.7} & \textbf{33.9} & \textbf{43.7} & \textbf{29.8} &	\textbf{38.1} &	46.4  \\
\bottomrule
\end{tabular}
\label{tab:sgdet_all}}
\end{table*}

%% file: mR_sota_pcls_sgcls.tex
\begin{table*}[tb!]
\centering
\caption{Comparative results for SGG tasks: PREDCLS and SGCLS, on AG \cite{Action_genome_Ji_2020_CVPR}, in terms of mean-Recall@K. Best results are shown in bold.}
\vspace{-0.5em}
\setlength{\tabcolsep}{12pt}
\resizebox{1\linewidth}{!}{%
\begin{tabular}{ccccccccccccc}
\toprule
  & \multicolumn{6}{c}{With Constraint} & \multicolumn{6}{c}{No Constraints}\\
  \cmidrule(lr){2-7} \cmidrule(lr){8-13} 
  Method & 
 \multicolumn{3}{c}{PredCLS} & \multicolumn{3}{c}{SGCLS} & \multicolumn{3}{c}{PredCLS} & \multicolumn{3}{c}{SGCLS} \\ 
    \cmidrule(lr){2-4} \cmidrule(lr){5-7}  \cmidrule(lr){8-10} \cmidrule(lr){11-13}  
  &mR@10 &mR@20 &mR@50 &mR@10 &mR@20 &mR@50 &mR@10  &mR@20 &mR@50 &mR@10 &mR@20 &mR@50 
\\
\midrule
\midrule
  RelDN~\cite{zhang2019vrd} & 6.2 & 6.2 & 6.2 & 3.4 & 3.4 & 3.4 & 31.2 & 63.1 & 75.5 & 18.6 & 36.9 & 42.6\\
  TRACE\cite{teng2021target}& 15.2& 15.2&15.2& 8.9& 8.9& 8.9 & 50.9&73.6& 82.7 &31.9& 42.7& 46.3\\
  STTran\cite{STTran_2021}& 37.8 & 40.1 & 40.2 & 27.2 & 28.0 & 28.0 & 51.4 & 67.7 & 82.7 & 40.7 & 50.1 & 58.8  \\ 
  STTran-TPI \cite{TPI_MM_22} & 37.3 & 40.6 & 40.6 & 28.3 & 29.3 & 29.3 & - & - & - & - & - & - \\
  TEMPURA&  \textbf{42.9}& \textbf{46.3}&\textbf{ 46.3} & \textbf{34.0} & \textbf{35.2 }& \textbf{35.2}  & \textbf{61.5} &	\textbf{85.1} & \textbf{98.0} &\textbf{ 48.3} & \textbf{61.1} & \textbf{66.4}  \\
\bottomrule
\end{tabular}
\label{tab:mR_sota_predcls_sgcls}}
\end{table*}


%% file: R_sota_pcls_sgcls.tex
\begin{table*}[tb!]
\centering
\caption{Comparative results for SGG tasks: PREDCLS and SGCLS, on AG \cite{Action_genome_Ji_2020_CVPR}, in terms of Recall@K. Best results are shown in bold. }
\vspace{-0.5em}
\setlength{\tabcolsep}{12pt}
\resizebox{1\linewidth}{!}{%
\begin{tabular}{ccccccccccccc}
\toprule
  & \multicolumn{6}{c}{With Constraint} & \multicolumn{6}{c}{No Constraints}\\
  \cmidrule(lr){2-7} \cmidrule(lr){8-13} 
  Method & 
 \multicolumn{3}{c}{PredCLS} & \multicolumn{3}{c}{SGCLS} & \multicolumn{3}{c}{PredCLS} & \multicolumn{3}{c}{SGCLS} \\ 
    \cmidrule(lr){2-4} \cmidrule(lr){5-7}  \cmidrule(lr){8-10} \cmidrule(lr){11-13}  
  &R@10 &R@20 &R@50 &R@10 &R@20 &R@50 &R@10  &R@20 &R@50 &R@10 &R@20 &R@50
\\
\midrule
\midrule


RelDN~\cite{zhang2019vrd} & 20.3 & 20.3 & 20.3 & 11.0 & 11.0 & 11.0 & 44.2 & 75.4 & 89.2 & 25.0 & 41.9 & 47.9 \\
  
TRACE \cite{teng2021target}& 27.5 & 27.5 & 27.5 & 14.8 & 14.8 & 14.8  & 72.6 & 91.6 &96.4 &37.1 &	46.7 &	50.5 \\

STTran \cite{STTran_2021}& 68.6 & 71.8 & 71.8 & 46.4	& 47.5 &	47.5  & 77.9 &	94.2 & 99.1 & 54.0 & 63.7 &66.4 \\

STTran-TPI \cite{TPI_MM_22}& \textbf{69.7} & 72.6 & 72.6 & \textbf{47.2} & 48.3 & 48.3 & - & - & - & - & - & - \\

APT \cite{APT_Li_2022_CVPR}& 69.4 & \textbf{73.8} & \textbf{73.8} & \textbf{47.2}	& \textbf{48.9} & \textbf{48.9}  & 78.5 & \textbf{95.1}&	99.2& 55.1 &	\textbf{65.1} &	\textbf{68.7}  \\

TEMPURA &  68.8 &	71.5 &	71.5 & \textbf{47.2} &	48.3&	48.3 & \textbf{80.4} &	94.2 &	\textbf{99.4} & \textbf{56.3} &	64.7 &	67.9  \\
\bottomrule
\end{tabular}
\label{tab:R_sota_predcls_sgcls}}
\vspace{-1em}
\end{table*}

  




%% file: mR_per_class.tex
\begin{figure}[htbp]
\centering
\begin{subfigure}[b]{0.3\linewidth}
  \centering
  \includegraphics[width=1\textwidth]{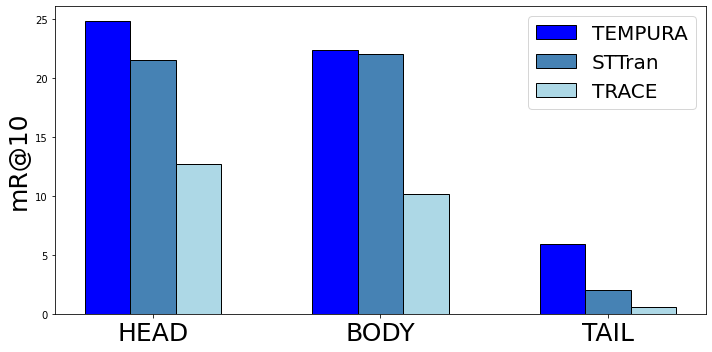}  
  \caption{SGDET}
\end{subfigure}
\begin{subfigure}[b]{0.3\linewidth}
  \centering
  \includegraphics[width=1\textwidth]{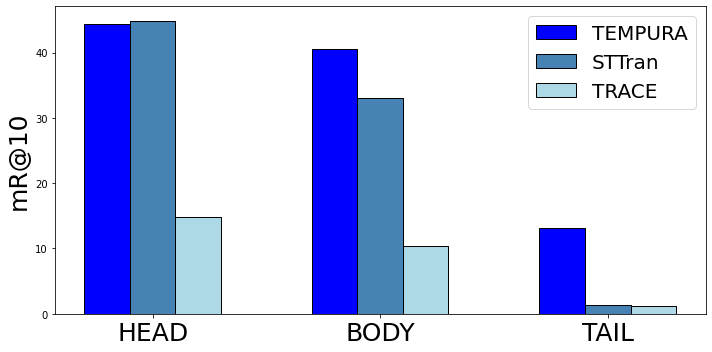}  
  \caption{SGCLS}
\end{subfigure}
\begin{subfigure}[b]{0.3\linewidth}
  \centering
  \includegraphics[width=1\textwidth]{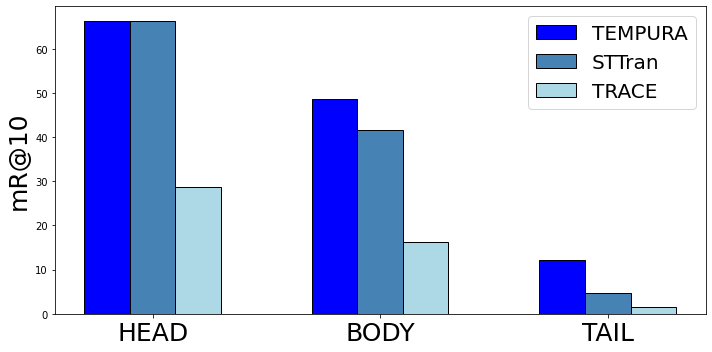}
  \caption{PREDCLS}
\end{subfigure}
\vspace{-0.5em}
\caption{Comparison of mR@10 for the HEAD, BODY and TAIL classes in Action Genome \cite{Action_genome_Ji_2020_CVPR} under the "with constraint" setup.}
  \vspace{-0.5em}
\label{fig:per_class_mR}
\end{figure}  

%% file: predcls_per_class_line.tex
\begin{figure}[ht]
\centering
\includegraphics[width=0.2\textwidth]{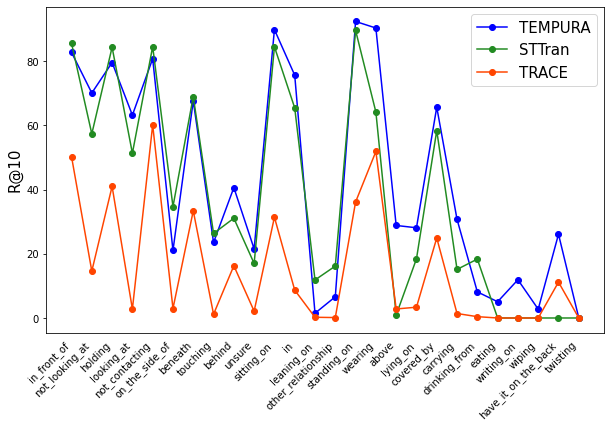} 
\vspace{-0.7em}
  \caption{\small 
Comparative per class performance for PREDCLS task. Results are in terms of R@10 under ``with constraint''. }
  \label{fig:per_class_line}
  \vspace{-1.6em}
\end{figure}

%% file: qualitative_results.tex
\begin{figure*}[ht]
\begin{center}
\includegraphics[width=0.93\linewidth]{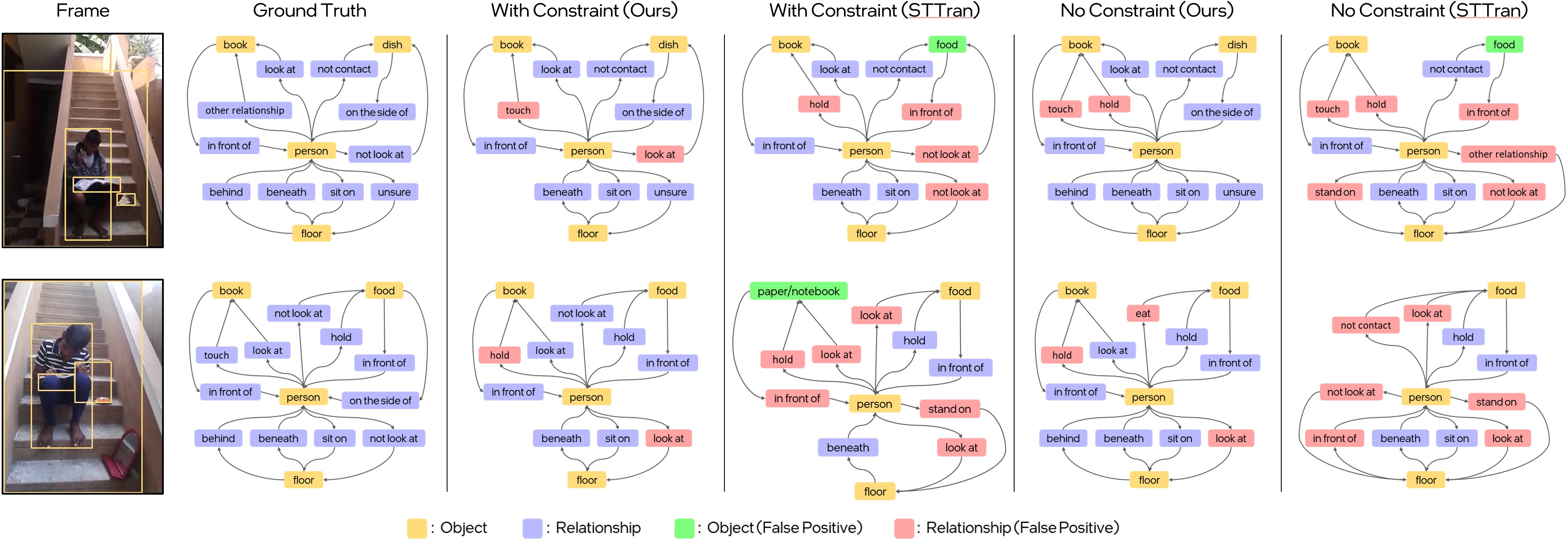}
\end{center}
\vspace{-1.5em}
  \caption{\small {\textbf{Comparative qualitative results}. From left to right: input video frames, ground truth scene graphs, scene graphs generated by TEMPURA, and the scene graphs generated by the baseline STTran \cite{STTran_2021}. Incorrect object and predicate predictions are shown in green and pink, respectively.}  }
\label{fig:qualitative}
\end{figure*}

%% file: sgcls_sgdet_ablations.tex
\begin{table*}[tb!]
\centering
\caption{Importance of uncertainty attenuation, memory guided debiasing, and temporally consistent object classification for SGCLS and SGDET.}
\vspace{-0.75em}
\setlength{\tabcolsep}{12pt}
\resizebox{1\linewidth}{!}{%
\begin{tabular}{ccc cccc cccc}
\toprule
  &  & & \multicolumn{4}{c}{With Constraint} & \multicolumn{4}{c}{No Constraints}\\
  \cmidrule(lr){4-7} \cmidrule(lr){8-11}
   \multirow{-2}{*}{\begin{tabular}[c]{@{}c@{}}Uncertainty \\ Attenuation\end{tabular}} & \multirow{-2}{*}{\begin{tabular}[c]{@{}c@{}}Memory guided \\ Debiasing\end{tabular}} & \multirow{-2}{*}{\begin{tabular}[c]{@{}c@{}}Temporal \\ Consistency\end{tabular}} &
  \multicolumn{2}{c}{SGCls} & \multicolumn{2}{c}{SGDet} & \multicolumn{2}{c}{SGCls} & \multicolumn{2}{c}{SGDet} \\ \cmidrule(lr){4-5} \cmidrule(lr){6-7} \cmidrule(lr){8-9} \cmidrule(lr){10-11} 
  & & & mR@10  & mR@20  & mR@10 & mR@20  & mR@10  & mR@20  & mR@10 & mR@20   
\\
\midrule
\midrule
-  & -  & -    & 27.2                                 & 28.0                                 & 16.5          & 20.8          & 40.7          & 50.1          & 20.9          & 29.7          \\ \hline
\checkmark                                                            &  -          & \checkmark                                                        & 30.6                                 & 31.9                                 & 16.7          & 21.1          & 43.5          & 58.9          & 20.9          & 30.5          \\
-   & \checkmark    & \checkmark                                                        & 31.8                                 & 33.2                                 & 16.8          & 20.9          & 45.7          & 59.7          & 21.7          & 30.7          \\
\checkmark   & \checkmark   &  -  & 30.9                                 & 32.1                                 & 17.0         & 21.4          & 45.7          & 59.3          & 21.6          & 30.1          \\
\checkmark                                                            & \checkmark                                                                & \checkmark                                                        & {\color[HTML]{000000} \textbf{34.0}} & {\color[HTML]{000000} \textbf{35.2}} & \textbf{18.5} & \textbf{22.6} & \textbf{48.3} & \textbf{61.1} & \textbf{24.7} & \textbf{33.9} \\
\bottomrule
\end{tabular}
\label{tab:sgcls_sgdet_ablations}}
\vspace{-1.0em}
\end{table*}

%% file: K_ablation.tex
\begin{table}[ht]
\centering
\caption{Performance of TEMPURA for varying numbers of GMM components, $\mathcal{K}$. Results are in terms of mR@10 for the \textbf{With Constraint} setup, with the best results shown in bold.
}
\setlength{\tabcolsep}{12pt}
\resizebox{1\linewidth}{!}{%
\begin{tabular}{@{}cccccc@{}}
\toprule
 \backslashbox{Task}{$\mathcal{K}$}& \multicolumn{1}{c}{1} & \multicolumn{1}{c}{2} & \multicolumn{1}{c}{4} & \multicolumn{1}{c}{6} & \multicolumn{1}{c}{8} \\ \midrule
\textit{PREDCLS}                                        & 40.1                  & 40.8                  & 42.6                  & \textbf{42.9}         & 42.1                  \\
\textit{SGCLS}                                          & 31.0                  & 33.1                  & \textbf{34.0}         & 32.7                  & 32.6                  \\
\textit{SGDET}                                          & 16.7                  & 17.0                  & \textbf{18.5}         & 18.2                  & 17.6                  \\ \bottomrule
\end{tabular}
}
\label{tab:K_ablation}
\vspace{-1.5em}
\end{table}

%% file: 5_conclusion.tex
\section{Conclusions}

The difficulty in generating dynamic scene graphs from videos can be attributed to several factors ranging from imbalanced predicate class distribution, video dynamics, temporal fluctuation of predictions, etc. Existing methods on dynamic SGG have mostly focused only on 
achieving high recall values, which are known to be biased towards head classes. In this work, we identify and address these sources of bias and propose a method, namely \textbf{TEMPURA}: \textbf{TE}mporal consistency and \textbf{M}emory \textbf{P}rototype
guided \textbf{U}nceRtainty \textbf{A}ttentuation for dynamic SGG that can compensate for those biases. We show that TEMPURA significantly outperforms existing methods in terms of mean recall metric, showing its efficacy in long-term unbiased visual relationship learning from videos. 

\noindent\textbf{Acknowledgments.} SN, KM, and ST were supported by Intel Corporation. SN and ARC were partially supported by ONR grant N000141912264 and NSF grant 1901379.

%% file: Supp.tex
\appendix




\noindent The supplementary material provides more details, results, and visualizations to support the main paper. In summary, we include additional implementation details, more experiments and ablation studies, analysis of our results, more qualitative visualizations, and a discussion on future works.

\section{Additional Implementation Details}

\paragraph{Predicate Class Distribution.} We define the \textit{HEAD}, \textit{BODY} and \textit{TAIL} relationship classes in Action Genome (Action Genome) \cite{Action_genome_Ji_2020_CVPR} as shown below,
\begin{itemize}
    \item \textit{HEAD} $\geq 100000$  training samples
    \item $8000$  training samples $ \leq $ \textit{BODY} $ < 100000 $ training samples
    \item \textit{TAIL} $< 8000 $ training samples
    
\end{itemize}

\paragraph{Experimental Setup.} The architecture of the PEG is kept the same as \cite{STTran_2021}. The Faster-RCNN object detector \cite{ren_frcnn} is first trained on Action Genome \cite{Action_genome_Ji_2020_CVPR} following \cite{STTran_2021,TPI_MM_22,APT_Li_2022_CVPR}. Following prior work per-class non-maximal
suppression at $0.4$ IoU is applied to reduce region proposals provided by the Faster-RCNN's RPN. The sequence encoder in the OSPU is designed with $3$ layers, each having $8$ heads for its multi-head attention. The dimension of its FFN projection is $1024$. During training, we reduce the initial learning rate by a factor of $0.5$ whenever the performance plateaus. All codes are run on a single NVIDIA RTX-3090.

\paragraph{Evaluation Metrics.} We follow the official implementation of \cite{VCTree_Tang_2019_CVPR} for the mean-Recall@K (mR@K) metric. Different from the standard Recall@K (R@K), mR@K is computed by first obtaining the  recall values of each predicate class and then averaging them over the total number of predicates. Therefore if an SGG model consistently fails to detect any of the visual relationships i.e., predicates, the mR@K value will drop considerably. This makes it a much more balanced metric compared to R@K, which is obtained by averaging recall values over the entire dataset. Therefore improvement on the high-frequent classes alone is sufficient for high R@K values. For all experiments, the reported results are in terms of image-based R@K and mR@K. Since the video-based R@K is a simple averaging of the per-frame measurements, most existing works adopt the image-based metrics \cite{STTran_2021,TPI_MM_22,APT_Li_2022_CVPR,ISG_arxiv22}. 

\paragraph{Baseline Performance.} For the baselines STTran \cite{STTran_2021}, TRACE \cite{teng2021target}, and RelDN \cite{RelDN}, we used their official code implementation to obtain the respective mR@K values. We obtained the mR@K values of STTran-TPI \cite{TPI_MM_22} from email discussions. The performance of HCRD supervised \cite{simple_baseline_bmvc}, and ISGG \cite{ISG_arxiv22} are taken from the reported values in \cite{ISG_arxiv22}. As explained in Section 4.1 of the main paper, SGG performance is typically evaluated under two different setups \textbf{With Constraint} and/or  \textbf{No Constraints}. All the baselines we compare with either follow the \textbf{With Constraint} \cite{TPI_MM_22,ISG_arxiv22,simple_baseline_bmvc} setup or the \textbf{No Constraints}  setup \cite{teng2021target} or both \cite{STTran_2021, APT_Li_2022_CVPR}.

\section{Additional Comparative Results}

\input{predcls_ablations.tex}

\input{L_intra_ablation.tex}
\input{mR_per_class_nc.tex}

\paragraph{Comparison of \textit{HEAD}, \textit{BODY} and \textit{TAIL} class performance, with SOTA, under No Constraints.} Comparative performance on the \textit{HEAD}, \textit{TAIL} and \textit{BODY} classes of Action Genome under the \textbf{No Constraints} setup are shown in Fig \ref{fig:per_class_mR_nc}. Similar to the \textbf{With Constraint} results shown in Fig $7$ of the main paper, TEMPURA outperforms both TRACE \cite{teng2021target} and STTran \cite{STTran_2021} in improving performance on the \textit{TAIL} and \textit{BODY} classes without significantly compromising performance on the \textit{HEAD} classes. While TRACE performs well under the \textbf{No Constraints} setup, its performance under the \textbf{With Constraints} setup is lacking (Fig 7 main paper). TEMPURA, on the other hand, shows consistent performance for both setups, beating TRACE and STTran  in generating more unbiased scene graphs. 

\section{Additional Ablations}

\paragraph{Ablations for PREDCLS task.} The impact of memory guided debiasing and uncertainty attenuation for the \textit{PREDCLS} task can be seen in Table \ref{tab:predcls_ablations}. Similar to the other SGG tasks (Table 4 of the main paper), incorporating both principles gives the best results. Since, for PREDCLS, the object bounding boxes and classes are already provided, the OSPU is inactive for this SGG task.

\paragraph{Comparison of \textit{HEAD}, \textit{BODY} and \textit{TAIL} class perforamnce for SGCLS and SGDET ablations.} From Fig \ref{fig:ablations_per_class_mR_wc} and Fig \ref{fig:ablations_per_class_mR_nc}, we can observe that using the full model (OSPU+MDU+GMM) gives the best performance for the \textit{BODY} and \textit{TAIL} classes for both SGCLS and SGDET tasks.

\paragraph{Impact of $\mathcal{L}_{intra}$.} The impact of $\mathcal{L}_{intra}$ can ascertained from Table \ref{tab:L_intra}. The results show that utilizing the intra-video contrastive loss $\mathcal{L}_{intra}$ boosts the sequence processing capability of the OSPU, leading to more consistent object classification and, consequently, more unbiased scene graphs.

\input{sgdet_sgcls_ablations_per_class_wc}
\input{sgdet_sgcls_ablations_per_class_nc}

\input{lambda_no_lambda.tex}
\input{lambda_ablation.tex}
\input{mR_per_class_lambda.tex}

\input{predictive_uncertainty_all.tex}

\paragraph{Ablations on $\lambda$.} The gradient scaling factor $\lambda$ regulates the influence of the direct PEG embedding $\bm{r}_{tem}^j$ of the $j^{th}$ subject-object pair and the compensatory information of the diffused memory feature $\bm{r}_{mem}^j$ as shown in Eq 11 in the main paper. $\lambda \in (0,1]$ i.e. $0 < \lambda \leq 1$. To obtain the optimal value of $\lambda$, we vary it within $[0.1,0.3,0.5,0.7,0.9]$ and observe the corresponding With Constraint R@10 and mR@10 values as shown in Fig \ref{fig:lambda_ablation}. As observed in Fig \ref{fig:lambda_ablation}, increasing the value of $\lambda$ causes the R@10 values to also increase before stagnating after a certain point. However, the corresponding mR@10 values start falling for higher $\lambda$ values. Since the Recall@K metric is an indicator of how well an SGG model is performing on the data-rich predicates, this indicates that for higher $\lambda$ values, the compensatory effect of $\bm{r}_{tem}^j$ is drastically reduced, and the PEG fails to generate more unbiased representations. On the other hand, if $\lambda$ is set to small values like $0.1$, high mR@10 values can be observed, but this comes at the expense of R@10 performance, indicating a drop in performance of the \textit{HEAD} classes due to excessive knowledge being transferred from these data-rich classes to the data-poor ones. Since the goal of unbiased SGG is not to perform well on the data-poor classes at the expense of data-rich classes, it is necessary to set $\lambda$ to an optimal value that gives the best balance between recall and mean-recall performance. As shown in Fig \ref{fig:lambda_ablation} the optimal $\lambda$ is $0.5$ for \textit{PREDCLS} and \textit{SGDET}, and $0.3$ for \textit{SGCLS}. From Fig \ref{fig:lambda_per_class} we can observe that setting $\lambda$ to the optimal values gives the best balance in performance over the \textit{HEAD}, \textit{BODY} and \textit{TAIL} classes as opposed to setting $\lambda$ to very low and very high values.

If $\lambda = 1$, there is no impact of $\bm{r}_{mem}^j$, and the model relies solely on the uncertainty attenuation of the GMM head. As explained in section 3.5 of the main paper,  $\bm{r}_{mem}^j$ essentially hallucinates information relevant to the data-poor classes otherwise missing from the original PEG embedding $\bm{r}_{tem}^j$ and the weighted residual operation of Eq 11 acts as a mechanism to diffuse this compensatory information back to $\bm{r}_{tem}^j$ in order to make it more balanced. Therefore, $\lambda$ can never be $0$; otherwise, there will be no PEG embedding to debias, and the MDU will never be able to teach the framework how to generate more unbiased embeddings. On the other hand, if  $\lambda$ is not used in the diffusion operation of Eq 11 i.e., if a standard non-weighted residual operation is used, the biased information from $\bm{r}_{tem}^j$ tends to overpower the effect of $\bm{r}_{mem}^j$. To verify this, we set up two experiments. In the first case, we set $\lambda =0$ and train the model, and in the second case,  we replace the weighted residual operation of Eq 11 with a simple residual operation i.e. $\bm{\hat{r}}_{tem}^{j} = \bm{r}_{tem}^j + \bm{r}_{mem}^j$ and then train the model. It can be observed from the With Constraint results shown in Table \ref{tab:lambda_no_lambda} that setting $\lambda$ to $0$ results in a significant drop in mR@K performance since the MDU is unable to diffuse the compensatory information back to the original PEG embedding rendering it ineffective in regularizing the model towards generating more unbiased predicate embeddings. Additionally, by comparing rows $2$ and $3$ in Table \ref{tab:lambda_no_lambda}, we can infer that utilization of $\lambda$ for the weighted residual operation of Eq 11 is necessary to get the best performance in terms of mean-recall.

\input{Role_of_MDU.tex}
\input{qualitative_results2.tex}

\section{Additional Analysis}

\paragraph{Are the effects of high uncertainty being attenuated?}
The predicate classification loss, 
$\mathcal{L}_p$ (Eq $16$) is designed to penalize the model if it predicts high uncertainty for any sample. This means the model progressively becomes more efficient in attenuating the effect of noisy samples, which inherently decreases its predictive uncertainty with the number of epochs. This can be visualized in Fig \ref{fig:predictive_uncertainty_all}, which shows the total predictive uncertainty of the full model for each SGG task. Both the epoch-specific \emph{aleatoric} and \emph{epistemic} uncertainties are obtained by averaging across all samples (subject-object pairs) over all classes.

\paragraph{Role of Memory Diffusion Unit.} As explained in section 3.5, the MDU and the predicate class-centric memory bank $\bm{\Omega}_R$ are used during the training phase as a structural meta-regularizer to debias the direct PEG embeddings and inherently teach the PEG how to learn more unbiased predicate embeddings. One might ask why the MDU and the training memory bank $\bm{\Omega}_R$ cannot be used as a network module to forward pass through during the inference like many memory-based works on long tail image recognition \cite{parisot2022long,zhu2020inflated}. This is because of the distributional shift between training and testing sets in the video SGG dataset. Such distributional shift also exist in standard image recognition datasets but is very minimal. That is not the case for video SGG data. For instance, unlike an image recognition dataset each sample of a visual relationship is not i.i.d. The visual relationship between a subject-object pair at each frame depends on the visual relationships (between the same pair) in the previous frames, and this temporal evolution is captured by $TempDec$ based on the motion information coming from the proposal features, shifting bounding boxes and union features of the subject-object pair. The temporal evolution of many visual relationships in the test videos can differ greatly from those in the training videos. This spatio-temporal information of each predicate class, \emph{in the entire training set}, is compressed into their respective memory prototypes $\bm{\omega}_p \in \bm{\Omega}_R$. Therefore utilizing these predicate memory prototypes (for the MDU operation) during inference biases the framework towards the training distribution which is antithetical to the purpose of the MDU. Additionally, the issue of triplet variability, shown in Fig 2 of the main paper, can further deepen the distribution shift since certain triplets associated with a relationship class can occur only during inference. For example, the triplets $< person-above-refrigerator >$ and $< person-lying \ on-bag >$ associated with the $above$ and $lying \ on$ predicates occur in only the test set videos of Action Genome \cite{Action_genome_Ji_2020_CVPR}. The information associated with these unique triplets is never incorporated  in  $\bm{\Omega}_R$, consequently impacting the predicate embedding if $\bm{\Omega}_R$ is used during inference which can lead to a drop in performance. We verify this by conducting a short experiment the results of which are shown in Table \ref{tab:role_of_mdu}.

\paragraph{More Qualitative Results.}
Some more qualitative results are shown in Fig \ref{fig:qualitative2}. It can be seen that TEMPURA prevents fewer false positives compared to the baseline STTran \cite{STTran_2021}.

\section{Limitations and Future Work}
 The progressive computation of the memory bank as a set of prototypical centroids does increase the training time, but as we showed in our results, this memory-guided training approach can result in more unbiased predicate representations that inherently help in the generation of more unbiased scene graphs.  In future works, we aim to explore a parallel memory computation approach that can perform at par with our current method. 



%% file: predcls_ablations.tex
\begin{table}[h]
\centering
\caption{Importance of uncertainty attenuation and memory guided meta-debiasing for PREDCLS.}
\setlength{\tabcolsep}{8pt}
\resizebox{1\linewidth}{!}{%
\begin{tabular}{cccccc}
\toprule
  &  &  \multicolumn{2}{c}{With Constraint} & \multicolumn{2}{c}{No Constraints}\\
  \cmidrule(lr){3-4} \cmidrule(lr){5-6}
  \multirow{-2}{*}{\begin{tabular}[c]{@{}c@{}}Uncertainty \\ Attenuation\end{tabular}} & \multirow{-2}{*}{\begin{tabular}[c]{@{}c@{}}Memory guided\\ Debiasing\end{tabular}} & 
    mR@10  & mR@20  & mR@10 & mR@20    \\
\midrule
\midrule
-                                                                           & -                                                                                & 37.8             & 40.1             & 51.4            & 67.7            \\ \midrule
\checkmark                                                   & -                                                                                & 40.2             & 44.0             & 55.1            & 77.3            \\
-                                                                           & \checkmark                                                        & 41.1             & 44.8             & 57.0            & 82.9            \\
\checkmark                                                   & \checkmark                                                        & \textbf{42.9}    & \textbf{46.3}    & \textbf{61.5}   & \textbf{85.1}   \\
\bottomrule
\end{tabular}
\label{tab:predcls_ablations}}

\end{table}

%% file: L_intra_ablation.tex
\begin{table}[tb!]
\centering
\caption{Impact of $\mathcal{L}_{intra}$. }
\setlength{\tabcolsep}{12pt}
\resizebox{1\linewidth}{!}{%
\begin{tabular}{ccccccccc}
\toprule
  &   \multicolumn{4}{c}{With Constraint} & \multicolumn{4}{c}{No Constraints}\\
  \cmidrule(lr){2-5} \cmidrule(lr){6-9} \\
   \multirow{-2}{*}{\begin{tabular}[c]{@{}c@{}}$\mathcal{L}_{intra}$\end{tabular}}   &
  \multicolumn{2}{c}{SGCls} & \multicolumn{2}{c}{SGDet}  & \multicolumn{2}{c}{SGCls} & \multicolumn{2}{c}{SGDet} \\ \cmidrule(lr){2-3} \cmidrule(lr){4-5} \cmidrule(lr){6-7} \cmidrule(lr){8-9} \\
  &  mR@10  & mR@20  & mR@10 & mR@20  & mR@10 & mR@20 & mR@10  & mR@20  
\\
\midrule
\midrule
-  &   32.1 & 33.2 & 17.9 & 22.1 & 46.5 & 60.4 & 23.5 & 32.8              \\
\checkmark &    \textbf{34.0} & \textbf{35.2} & \textbf{18.5} & \textbf{22.6}  & \textbf{48.3} & \textbf{61.1} & \textbf{24.7} & \textbf{33.9}   \\

\bottomrule
\end{tabular}
\label{tab:L_intra}
}
\end{table}

%% file: mR_per_class_nc.tex
\begin{figure*}[http]
\centering
\begin{subfigure}[b]{0.32\linewidth}
  \centering
  \includegraphics[width=1\textwidth]{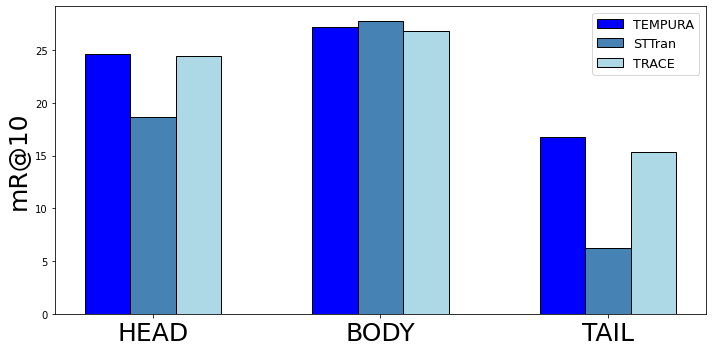}  
  \caption{SGDET}
\end{subfigure}
\begin{subfigure}[b]{0.32\linewidth}
  \centering
  \includegraphics[width=1\textwidth]{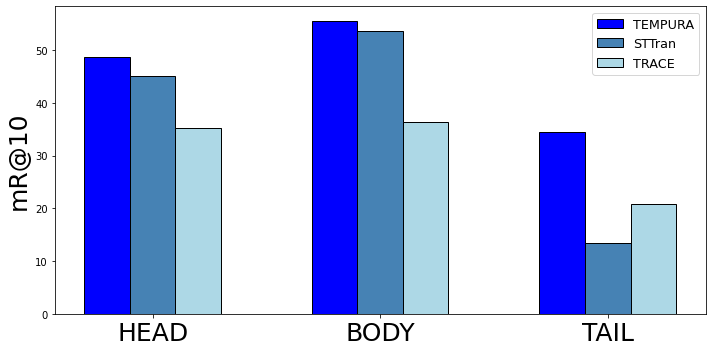}  
  \caption{SGCLS}
\end{subfigure}
\begin{subfigure}[b]{0.32\linewidth}
  \centering
  \includegraphics[width=1\textwidth]{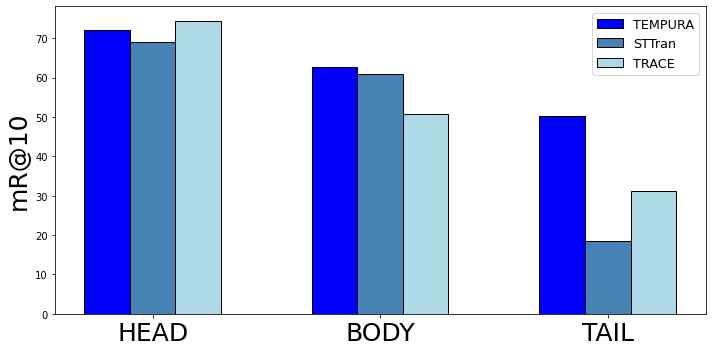}
  \caption{PREDCLS}
\end{subfigure}
\caption{Comparison of mR@10 for the HEAD, BODY and TAIL classes in Action Genome \cite{Action_genome_Ji_2020_CVPR} under the "No constraints" setup.}
\label{fig:per_class_mR_nc}
\end{figure*}  

%% file: sgdet_sgcls_ablations_per_class_wc.tex
\begin{figure}[t]
\centering
\begin{subfigure}[b]{0.45\linewidth}
  \centering
  \includegraphics[width=1\textwidth]{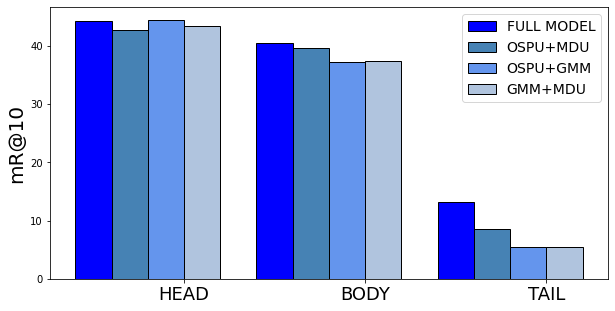}  
  \caption{SGCLS}
\end{subfigure}
\begin{subfigure}[b]{0.45\linewidth}
  \centering
  \includegraphics[width=1\textwidth]{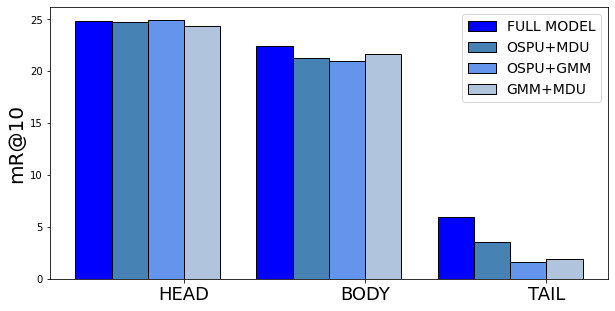}  
  \caption{SGDET}
\end{subfigure}
\caption{Comparison of mR@10 for the HEAD, BODY, and TAIL classes in Action Genome \cite{Action_genome_Ji_2020_CVPR} for different ablation setups of Table 4. Performances reported under the ”with constraint” setup.}
\label{fig:ablations_per_class_mR_wc}
\end{figure}  

%% file: sgdet_sgcls_ablations_per_class_nc.tex
\begin{figure}[t]
\centering
\begin{subfigure}[b]{0.45\linewidth}
  \centering
  \includegraphics[width=1\textwidth]{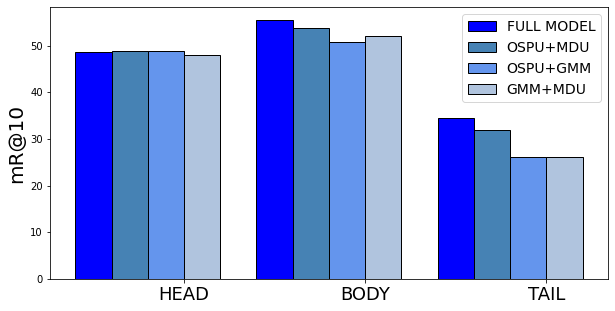}  
  \caption{SGCLS}
\end{subfigure}
\begin{subfigure}[b]{0.45\linewidth}
  \centering
  \includegraphics[width=1\textwidth]{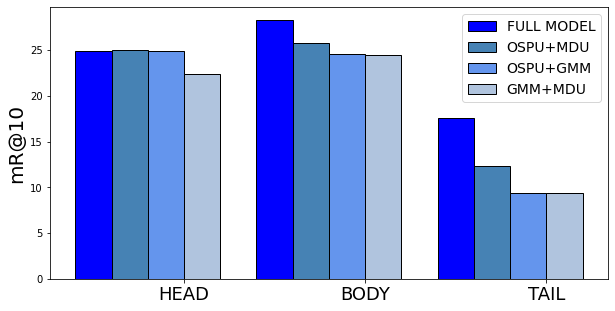}  
  \caption{SGDET}
\end{subfigure}
\caption{Comparison of mR@10 for the HEAD, BODY, and TAIL classes in Action Genome \cite{Action_genome_Ji_2020_CVPR} for different ablation setups of Table 4. Performances reported under the ”no constraints” setup.}
\label{fig:ablations_per_class_mR_nc}
\end{figure}

%% file: lambda_no_lambda.tex
\begin{table}[tb!]
\centering
\caption{Comparitive performance of TEMPURA for three different settings of $\lambda$. No $\lambda$ corresponds to when the weighted residual operation of Eq 11 is replaced with a standard residual connection. The optimal values of $\lambda$ are $0.5, \ 0.3$, and $0.5$ for PREDCLS, SGCLS, and SGDET, respectively.}
\setlength{\tabcolsep}{12pt}
\resizebox{1\linewidth}{!}{%
\begin{tabular}{ccc cccc}
\toprule

  \multirow{2}{*}{$\lambda$ Setting} & \multicolumn{2}{c}{PredCLS} & \multicolumn{2}{c}{SGCls} & \multicolumn{2}{c}{SGDET}  \\ \cmidrule(lr){2-3} \cmidrule(lr){4-5} \cmidrule(lr){6-7} 
   &  mR@10  & mR@20  & mR@10 & mR@20  & mR@10  & mR@20   
\\
\midrule
\midrule
$\lambda = 0$  & 31.7 & 36.9 & 25.0 & 26.2 & 13.4 & 17.9 \\
No $\lambda$ & 39.4 & 43.1 & 30.8 & 32.2 & 17.3 & 21.6 \\
Optimal $\lambda$ & \textbf{42.9} & \textbf{46.3} &\textbf{ 34.0} & \textbf{35.2} & \textbf{18.5} & \textbf{22.6} \\
\bottomrule
\end{tabular}
\label{tab:lambda_no_lambda}
}
\end{table}

%% file: lambda_ablation.tex
\begin{figure*}[http]
\centering
\begin{subfigure}[b]{0.3\linewidth}
  \centering
  \includegraphics[width=1\textwidth]{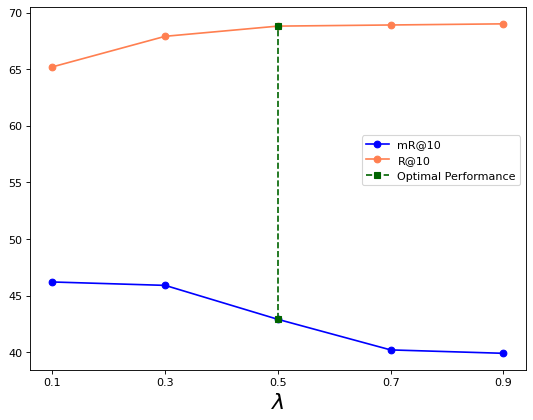}  
  \caption{PREDCLS}
\end{subfigure}
\begin{subfigure}[b]{0.305\linewidth}
  \centering
  \includegraphics[width=1\textwidth]{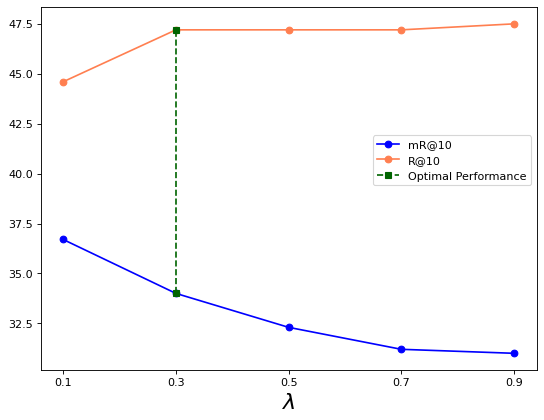}  
  \caption{SGCLS}
\end{subfigure}
\begin{subfigure}[b]{0.3\linewidth}
  \centering
  \includegraphics[width=1\textwidth]{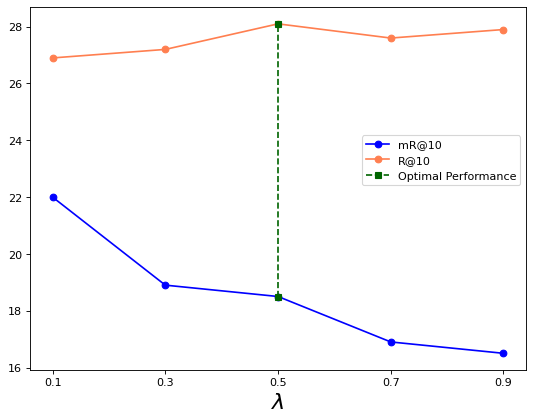}
  \caption{SGDET}
\end{subfigure}
\caption{Comparison of R@10 and mR@10 performance of TEMPURA for different values of $\lambda$.}
\label{fig:lambda_ablation}
\end{figure*}  

%% file: mR_per_class_lambda.tex
\begin{figure*}[http]
\centering
\begin{subfigure}[b]{0.32\linewidth}
  \centering
  \includegraphics[width=1\textwidth]{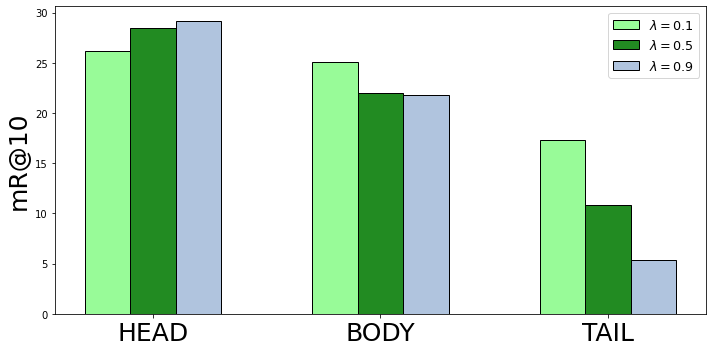}  
  \caption{SGDET}
\end{subfigure}
\begin{subfigure}[b]{0.32\linewidth}
  \centering
  \includegraphics[width=1\textwidth]{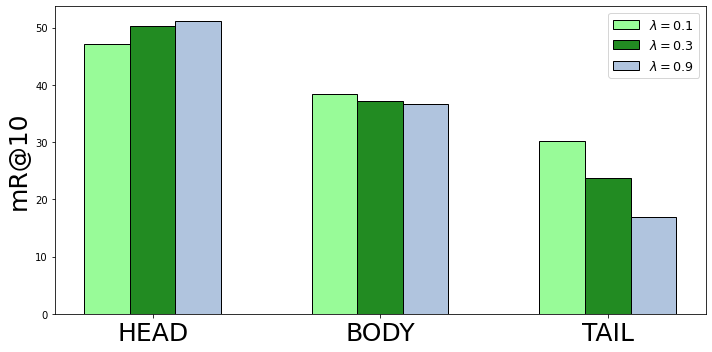}  
  \caption{SGCLS}
\end{subfigure}
\begin{subfigure}[b]{0.32\linewidth}
  \centering
  \includegraphics[width=1\textwidth]{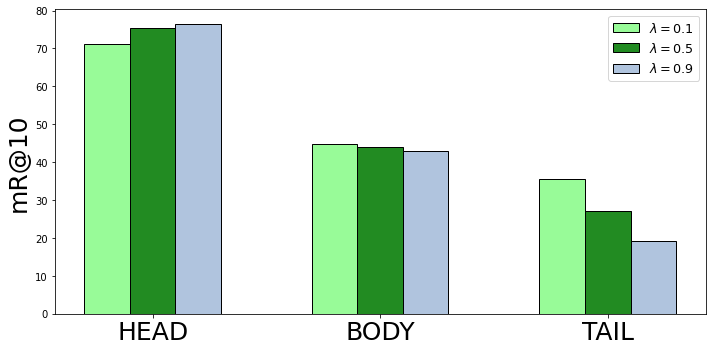}
  \caption{PREDCLS}
\end{subfigure}
\caption{Comparison of mR@10 for the HEAD, BODY and TAIL classes in Action Genome \cite{Action_genome_Ji_2020_CVPR} for different $\lambda$ values.}
  \vspace{-1em}
\label{fig:lambda_per_class}
\end{figure*}  

%% file: predictive_uncertainty_all.tex
\begin{figure}[tp]
\centering
\begin{subfigure}[b]{0.45\linewidth}
  \centering
  \includegraphics[width=0.9\textwidth]{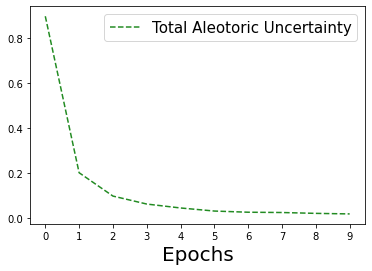}  
  \caption{}
\end{subfigure}
\begin{subfigure}[b]{0.46\linewidth}
  \centering
  \includegraphics[width=0.9\textwidth]{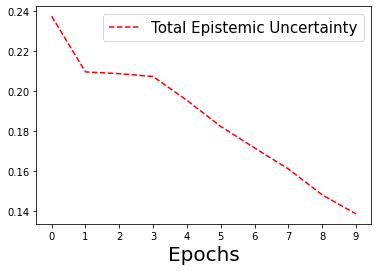}
  \caption{}
\end{subfigure}
\begin{subfigure}[b]{0.45\linewidth}
  \centering
  \includegraphics[width=0.9\textwidth]{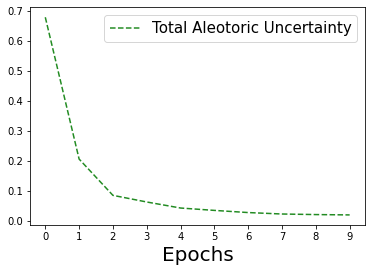}
  \caption{}
\end{subfigure}
\begin{subfigure}[b]{0.46\linewidth}
  \centering
  \includegraphics[width=0.9\textwidth]{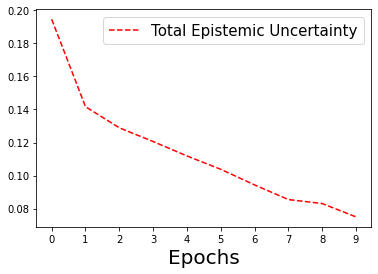}
  \caption{}
\end{subfigure}
\begin{subfigure}[b]{0.45\linewidth}
  \centering
  \includegraphics[width=0.9\textwidth]{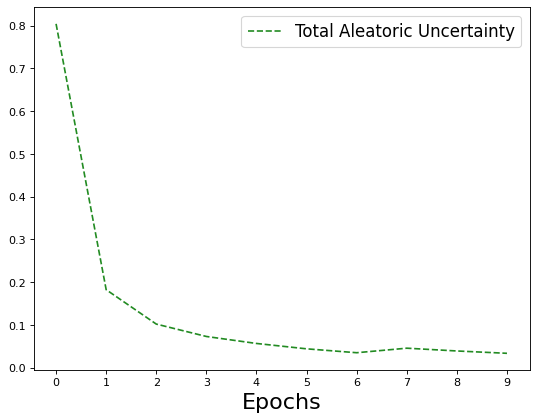}
  \caption{}
  \label{fig:sgdet_al_unc}
\end{subfigure}
\begin{subfigure}[b]{0.46\linewidth}
  \centering
  \includegraphics[width=0.9\textwidth]{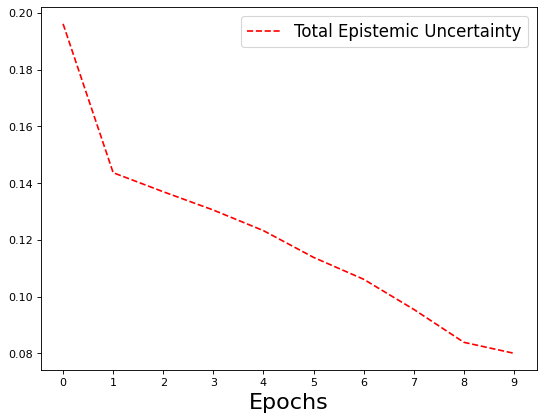}
  \caption{}
  \label{fig:sgdet_ep_unc}
\end{subfigure}
\caption{Top to Bottom: Predictive uncertainty for \textit{PREDCLS} (a,b), Predictive Uncertainty for \textit{SGCLS} (c,d) and Predictive Uncertainty for \textit{SGDET} (e,f).}
\vspace{-1em}
\label{fig:predictive_uncertainty_all}
\end{figure}

%% file: Role_of_MDU.tex
\begin{table*}[ht]
\centering
\caption{Comparison of performance when the memory bank $\bm{\Omega}_R$ and MDU is used and not used during inference. In both cases, the same model is used during inference which has been trained using the MDU. The results in the first row correspond to when MDU acts as a network module, and those in the second row correspond to when MDU is used as a meta-regularization unit which is its intended purpose. It can be observed from the results that incorporating the training memory bank $\bm{\Omega}_R$ for the test videos can bias the relationship representations towards the training distribution, defeating the purpose of the MDU. }
\setlength{\tabcolsep}{12pt}
\resizebox{1\linewidth}{!}{%
\begin{tabular}{ccccccccccccc}
\toprule
  &   \multicolumn{6}{c}{With Constraint} & \multicolumn{6}{c}{No Constraints}\\
  \cmidrule(lr){2-7} \cmidrule(lr){8-13} \\
   \multirow{-2}{*}{\begin{tabular}[c]{@{}c@{}}MDU used \\ during Inference\end{tabular}}  & \multicolumn{2}{c}{PredCLS} &
  \multicolumn{2}{c}{SGCls} & \multicolumn{2}{c}{SGDet} & \multicolumn{2}{c}{PredCLS} & \multicolumn{2}{c}{SGCls} & \multicolumn{2}{c}{SGDet} \\ \cmidrule(lr){2-3} \cmidrule(lr){4-5} \cmidrule(lr){6-7} \cmidrule(lr){8-9} \cmidrule(lr){10-11} \cmidrule(lr){12-13} \\
  &  mR@10  & mR@20  & mR@10 & mR@20  & mR@10 & mR@20 & mR@10  & mR@20  & mR@10 & mR@20  & mR@10 & mR@20
\\
\midrule
\midrule
\checkmark  &   38.5 & 42.0 & 29.9 & 31.2 & 16.1 & 20.4 & 53.6 & 80.0 & 42.5 & 57.5 & 19.4 & 29.6              \\
- &   \textbf{42.9}    & \textbf{46.3} & \textbf{34.0} & \textbf{35.2} & \textbf{18.5} & \textbf{22.6}  & \textbf{61.5}   & \textbf{85.1} & \textbf{48.3} & \textbf{61.1} & \textbf{24.7} & \textbf{33.9}   \\

\bottomrule
\end{tabular}
\label{tab:role_of_mdu}
}
\end{table*}

%% file: qualitative_results2.tex
\begin{figure*}[ht]
\begin{center}
\includegraphics[width=0.9999\linewidth]{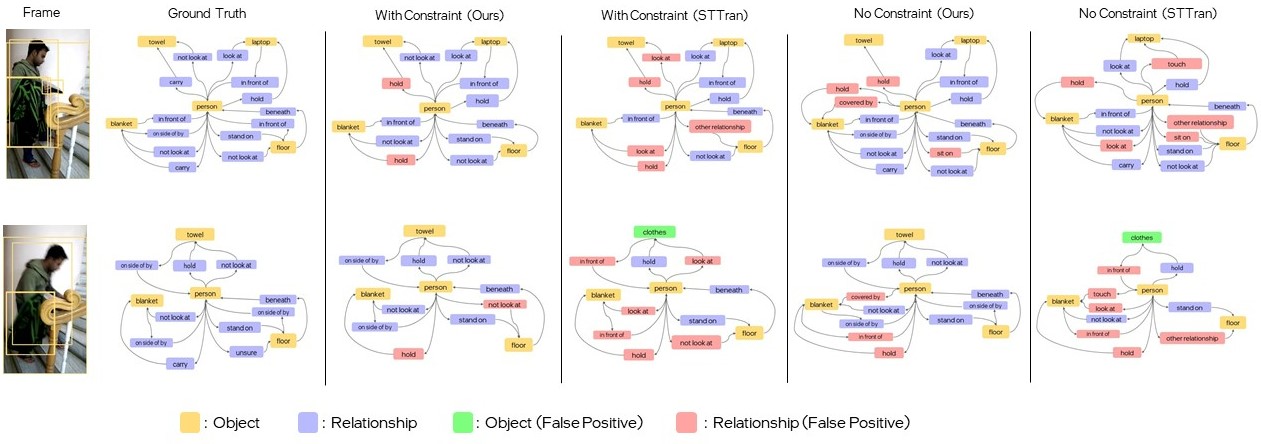}
\end{center}
  \caption{\small {\textbf{Comparative qualitative results}. From left to right: input video frames, ground truth scene graphs, scene graphs generated by TEMPURA and the scene graphs generated by the baseline STTran. Incorrect object and predicate predictions are shown in green and pink, respectively.}  }
\label{fig:qualitative2}
\vspace{-1em}
\end{figure*}